\documentclass{bmvc2k}

\usepackage[utf8]{inputenc} 
\usepackage[T1]{fontenc}    
\usepackage{hyperref}       
\usepackage{url}            
\usepackage{booktabs}       
\usepackage{amsfonts}       
\usepackage{nicefrac}       
\usepackage{microtype}      
\usepackage{xcolor}         
\usepackage{soul}           

\usepackage{float}
\usepackage{algorithm}
\usepackage[noend]{algpseudocode}
\usepackage{multicol}
\usepackage{amssymb}
\usepackage{amsmath}
\usepackage{graphicx}
\usepackage{threeparttable}
\usepackage{multirow}
\usepackage{booktabs, makecell}
\usepackage{placeins}
\usepackage{needspace}
\usepackage{subcaption}

\newcommand*{\Scale}[2][4]{\scalebox{#1}{$#2$}}%
\usepackage{caption}                  
\title{Audio-Guided Visual Editing with Complex Multi-Modal Prompts}

\addauthor{Hyeonyu Kim}{hykim@maum.ai}{1}
\addauthor{Seokhoon Jeong}{shjd0246@unist.ac.kr}{2}
\addauthor{Seonghee Han}{seonghee@unist.ac.kr}{2}
\addauthor{Chanhyuk Choi}{chan4184@unist.ac.kr}{2}
\addauthor{Taehwan Kim}{taehwankim@unist.ac.kr}{2}

\addinstitution{
 MAUM AI Inc.\\
 Republic of Korea
}
\addinstitution{
 Artificial Intelligence Graduate School\\
 UNIST \\
 Republic of Korea
}

\runninghead{
    \footnotesize{Kim et al}
}
{
    \footnotesize{Audio-Guided Visual Editing with Complex Multi-Modal Prompts}
}

\begin{document}

\maketitle
\begin{abstract}

Visual editing with diffusion models has made significant progress but often struggles with complex scenarios that textual guidance alone could not adequately describe, highlighting the need for additional non-text editing prompts. In this work, we introduce a novel audio-guided visual editing framework that can handle complex editing tasks with multiple text and audio prompts without requiring additional training. Existing audio-guided visual editing methods often necessitate training on specific datasets to align audio with text, limiting their generalization to real-world situations. We leverage a pre-trained multi-modal encoder with strong zero-shot capabilities and integrate diverse audio into visual editing tasks, by alleviating the discrepancy between the audio encoder space and the diffusion model's prompt encoder space. Additionally, we propose a novel approach to handle complex scenarios with multiple and multi-modal editing prompts through our separate noise branching and adaptive patch selection. Our comprehensive experiments on diverse editing tasks demonstrate that our framework excels in handling complicated editing scenarios by incorporating rich information from audio, where text-only approaches fail.

\end{abstract}

\vspace{-0.5cm}
\section{Introduction} \label{sec:intro}

Recent advancements in text-to-image diffusion models have facilitated visual editing tasks, which require precise manipulation of images or videos while preserving essential elements \cite{hertz2022prompt, mokady2023null, tumanyan2023plug, geyer2023tokenflow}. However, relying solely on text prompts limits their effectiveness in complex editing scenarios where descriptions may be ambiguous or insufficient \cite{kumari2023multi, brack2024ledits++, zhao2024uni, he2024flexecontrol}. To address this limitation, researchers have incorporated non-text conditions—such as edge map, segmentation mask, and depth map—to provide richer and more precise control over the editing process \cite{mo2024freecontrol, zhao2024uni, he2024flexecontrol, lin2024ctrl}. Audio is one of the widely used modalities and can provide rich and dynamic information, thus being effective for visual editing. However, in audio-guided visual editing, existing methods often require additional training on specific datasets to align audio with the text \cite{yang2023align, biner2024sonicdiffusion, li2022learning}, limiting their generalization to diverse real-world scenarios.

\begin{figure}[t!]
    \centering
    \includegraphics[trim=0cm 0.5cm 0cm 0cm , width= 0.9\textwidth, clip]{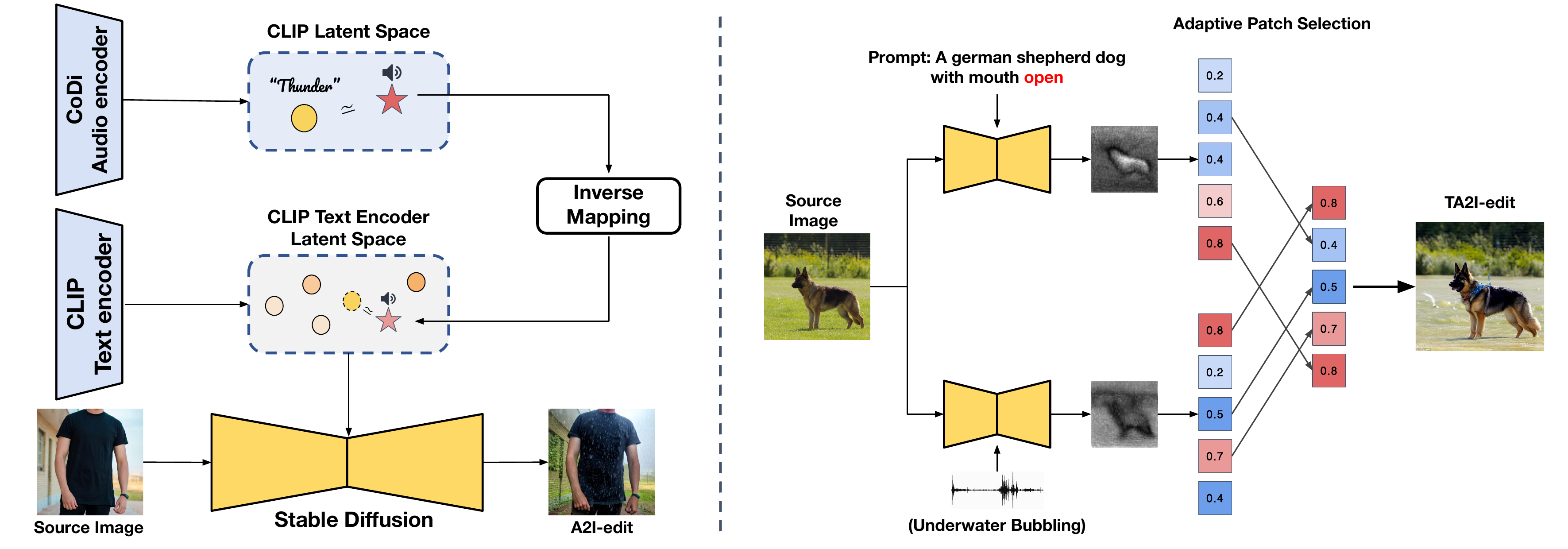}
    \caption{\textbf{Left}: To address the limited generalization of previous methods that require additional training to align audio data with text, our framework leverages an aligned multimodal encoder pretrained on large-scale datasets without extra training. \textbf{Right}: Instead of enhancing diffusion models to handle multiple editing prompts through additional training, we propose a novel method, separate noise branching and adaptive patch selection, which can effectively combine multiple and multi-modal prompts, while preserving each editing effect.}
    \label{fig:model_architecture}
\end{figure}

We address this limitation by leveraging a pre-trained aligned multi-modal encoder trained on large-scale text-audio and audio-visual datasets \cite{tang2023any}, which demonstrates strong zero-shot capabilities. However, integrating this encoder with diffusion models such as Stable Diffusion \cite{rombach2022high} faces challenges due to the discrepancy between the encoder's output space and the CLIP text encoder space used in Stable Diffusion, as illustrated in Figure \ref{fig:model_architecture}. We overcome this incompatibility by introducing a simple mapping technique involving only a matrix inversion, effectively incorporating audio features in visual editing without extensive retraining.

Achieving richer and more complex visual editing often requires integrating multiple editing prompts \cite{mo2024freecontrol, zhao2024uni, he2024flexecontrol, lin2024ctrl}, such as combining audio and text prompts. Existing approaches typically enhance diffusion models with additional modules to handle multiple editing prompts, requiring further training. In contrast, we propose a novel separate noise branching and adaptive patch selection as shown in Figure \ref{fig:model_architecture}, which can effectively integrate multiple editing signals without training.

To comprehensively evaluate our framework, we construct new benchmark datasets, \textit{PIEBench-multi} and \textit{DAVIS-multi}, by extending existing text-guided visual editing benchmarks to include audio editing prompts. Through comprehensive experiments and analysis, we demonstrate that our framework effectively handles complex editing scenarios by effectively incorporating rich information from audio, which existing text-only methods struggle with. We will publicly release our benchmarks to promote subsequent research. 

Our contributions can be summarized as follows:

\begin{itemize} 
    \item We present a novel zero-shot approach to integrate audio into visual editing, which delivers strong performance across diverse types of audio by effectively integrating a pre-trained aligned multi-modal encoder without any further training.
    
    \item We propose a novel approach of separate noise branching and adaptive patch selection, which can handle complex visual editing scenarios with multiple editing prompts without further training.
    
    \item Comprehensive experiments and analysis on our new benchmark datasets, \textit{PIEBench-multi} and \textit{DAVIS-multi}, demonstrate that our framework effectively handles complex editing scenarios involving both text and audio prompts, where existing text-only methods struggle. We will also release our benchmarks to promote subsequent research. 
\end{itemize}

\vspace{-25pt}

\section{Related Work}\label{sec:related}

\subsection{Diffusion Models in Multi-modal Generation}
With the groundbreaking achievements in text-to-image diffusion models \cite{rombach2022high, ramesh2022hierarchical, saharia2022photorealistic, podell2023sdxl}, researchers have actively explored the application of diffusion models in various output modalities, including text \cite{gong2022diffuseq, wu2024ar, luo2024diff}, audio \cite{popov2021grad, liu2023audioldm, shen2023naturalspeech} and video \cite{khachatryan2023text2video, blattmann2023align, luo2023videofusion}. Furthermore, incorporating non-text conditions, such as video-to-audio \cite{du2023conditional, luo2024diff}, image-to-video \cite{gong2024atomovideo, ni2023conditional}, and audio-to-image \cite{lee2023generating, yang2023align} has led to impressive results. Notably, the integration of visual, textual, and auditory information as either condition or output \cite{tang2023any, wu2023next, xing2024seeing} enables us to handle a broader range of tasks within a unified framework. 

\subsection{Diffusion Models in Editing}
Building on the observation that DDIM sampling can be reversed to reconstruct or interpolate visual data \cite{song2020denoising}, numerous studies have focused on diffusion-based editing. These works primarily target image editing with text prompts \cite{hertz2022prompt, mokady2023null, tumanyan2023plug, ju2023direct}, aiming to achieve faithful inversion for enhanced editing capabilities. Additional control mechanisms have been incorporated alongside DDIM inversion, such as cross-attention \cite{hertz2022prompt}, unconditional embeddings \cite{mokady2023null}, and spatial and self-attention features \cite{tumanyan2023plug}.

Text-guided video editing has also drawn significant interest \cite{bar2022text2live, wu2023tune, qi2023fatezero, geyer2023tokenflow}, with an emphasis on preserving temporal consistency across edited frames. Moreover, the field of diffusion-based editing has expanded to include sound, which is utilized either as a prompt \cite{yang2023align, lee2023soundini, biner2024sonicdiffusion, li2022learning, lin2025zero} or as the subject of editing \cite{liu2023audioldm, wang2024audit, manor2024zero}.

Recently, it has been recognized that text prompts alone are insufficient for handling complex editing scenarios \cite{kumari2023multi, brack2024ledits++, zhao2024uni, he2024flexecontrol}. To address this limitation, researchers have explored incorporating non-text conditions—such as edge map, segmentation mask, and depth map—to achieve richer and more precise control \cite{mo2024freecontrol, zhao2024uni, he2024flexecontrol, lin2024ctrl}. However, most of these approaches expand the capabilities of diffusion models by adding and training additional modules to the text-to-image diffusion model, which requires extra training with paired datasets. CoDi-2 \cite{tang2023codi} demonstrates instructional editing capability with multiple and multi-modal prompts, but requires extensive training on paired datasets of original data \cite{hertz2022prompt}. In contrast, our approach does not require training with paired data and has strong zero-shot editing ability.

\vspace{-15pt}
\section{Method}
\label{sec:method}

\subsection{Background} \label{sec:Method:background}
A family of Latent Diffusion Model \cite{rombach2022high, liu2023audioldm} transforms a high-dimensional input data $x$ into a lower-dimensional latent $\mathbf{z}$ using the encoder $\mathcal{E}$ and decoder $\mathcal{D}$, where $\hat{x} = \mathcal{D}(\mathcal{E}(x))$. Subsequently, a time-conditioned U-net, denoted as $\theta$, conducts a sequential diffusion process to predict $\mathbf{z}_0$ from random noise $\mathbf{z}_T$, considering the textual condition $\mathcal{C} = \psi(\mathcal{P})$ as Equation \ref{eq:ldm}, where $\psi$ refers to the prompt encoder and $\mathcal{P}$ is editing prompt.

\vspace{-5pt}
\begin{equation}
  \label{eq:ldm}
    \mathcal{L}_{ldm} = \mathbb{E}_{\mathcal{E}(x), \epsilon \sim \mathcal{N}(0,1), t} \lVert \epsilon - \epsilon_\theta (\mathbf{z}_t, t, \mathcal{C}) \lVert_{\Scale[0.7]{2}}^{\Scale[0.7]{2}},
\end{equation}

DDIM inversion \cite{song2020denoising} allows the original data to be encoded through an invertible transformation as expressed in Equation  \ref{eq:ddim_inv}, resulting in initial noise $\mathbf{z}^{\ast}_T$ capable of reconstruction and interpolation. In editing tasks, conditional DDIM inversion with textual description $\mathcal{P}_{\Scale[0.7]{inv}}$ is commonly employed, where the initial noise $\mathbf{z}^{\ast}_T$ is utilized along with the editing prompt $\mathcal{P}$ in sampling stage to maintain the structure and semantic layout. We use $\ast$ to distinguish the diffusion trajectory of DDIM inversion and editing. 

\vspace{-5pt}
\begin{equation}
  \label{eq:ddim_inv}
  \resizebox{0.65\columnwidth}{!}{$
    \mathbf{z}_{t+1}^{\ast} = \sqrt{\frac{\alpha_{t+1}}{\alpha_t}}\mathbf{z}_t^{\ast} + 
    \left( \sqrt{\frac{1}{\alpha_{t+1}} - 1} -
    \sqrt{\frac{1}{\alpha_t} - 1} \right) \cdot 
    \epsilon_\theta (\mathbf{z}_t^{\ast}, t, \mathcal{C}_{\text{inv}} )
  $}
\end{equation}

However, this approach often leads to undesired changes \cite{mokady2023null}, prompting the additional controls over the diffusion process for fine-grained editing \cite{couairon2022diffedit, meng2021sdedit, mokady2023null, tumanyan2023plug}. PnP-Diffusion \cite{tumanyan2023plug} has demonstrated the effectiveness of injecting spatial features and self-attention to preserve the original image content. Specifically, spatial features of the fourth decoder layer and the self-attention matrix are extracted during the DDIM inversion, which are then injected in the editing stage. TokenFlow \cite{geyer2023tokenflow}, which we adopt for video editing, adapts the PnP-Diffusion to video editing. It first randomly samples a subset of keyframes from the input video and jointly edits them via an extended self-attention mechanism as in Tune-A-Video \cite{wu2023tune}. It then propagates the edited tokens from each keyframe’s self-attention map to its neighboring frames to ensure temporal coherence.



\begin{figure*}[t]
    \centering
    \includegraphics[trim=0cm 0.2cm 0cm 0cm , width=0.7\linewidth, clip]{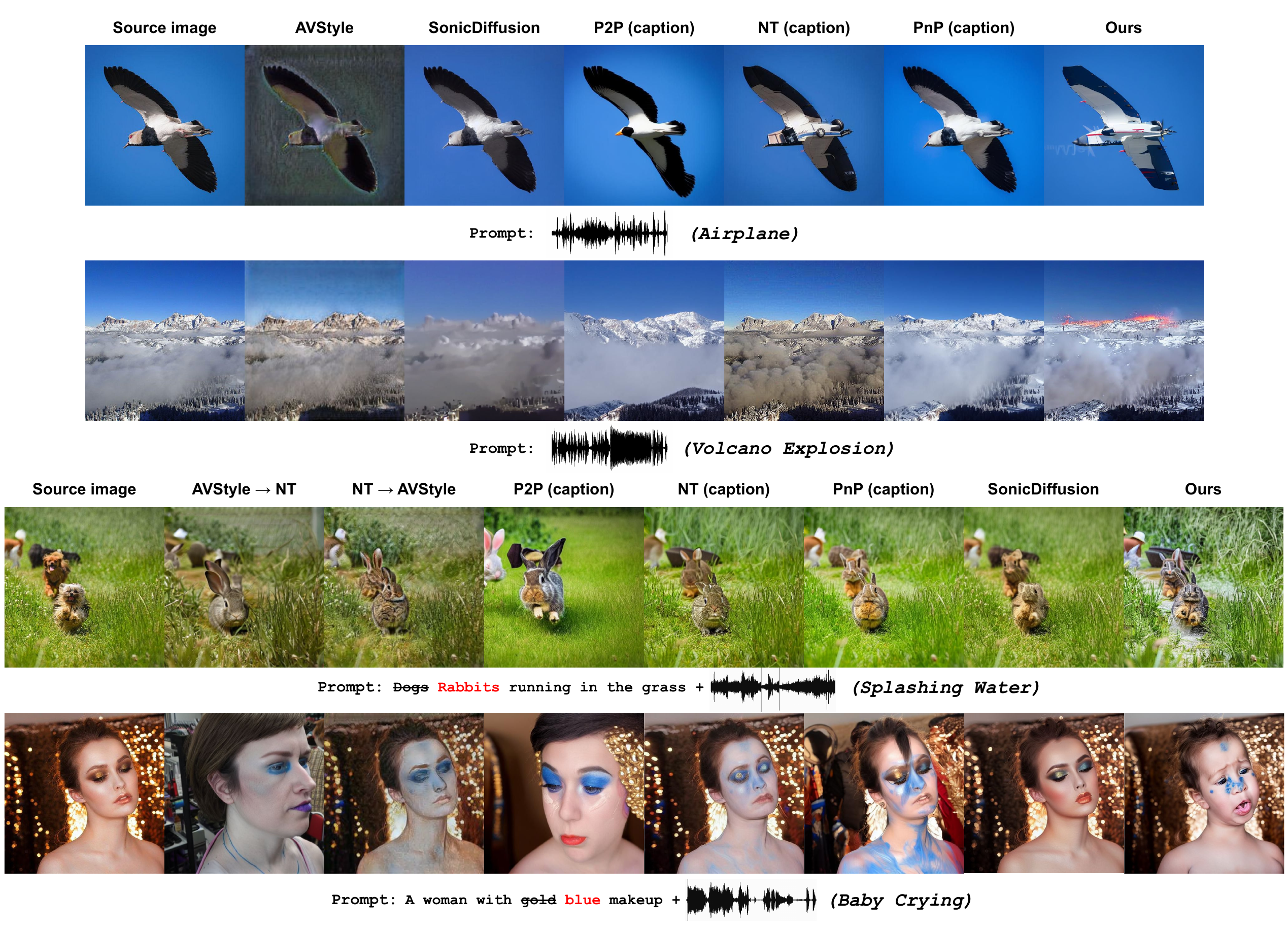}
    \caption{Editing results for Audio-guided Image Editing (\textbf{A2I-edit}) and Text and Audio-guided Image Editing (\textbf{TA2I-edit})}
    \label{fig:qualitative_image}
\end{figure*}

\subsection{Training-Free Integration of Audio Prompts}
\label{sec:Method:anytoedit:audio}

With the rapid advancement of visual editing, research has evolved to integrate more diverse conditions beyond simple text prompts \cite{mo2024freecontrol, zhao2024uni, he2024flexecontrol, lin2024ctrl, tang2023codi}.  
Most audio-conditioned approaches fine-tune the diffusion model to align sound with text \cite{yang2023align,biner2024sonicdiffusion,li2022learning}, but such adaptation couples the model to the specific training data and generalizes poorly to in-the-wild audio.  

To overcome this limitation, we adopt a large-scale multimodal encoder such as ImageBind \cite{girdhar2023imagebind} and CoDi \cite{tang2023any}. Specifically, we utilize CoDi because (i) it relies on the same OpenCLIP-L/14 backbone as Stable Diffusion 1.x, common in visual-editing pipelines, and (ii) it is trained on the largest public audio corpus, giving robust audio representations.

However, integrating CoDi with Stable Diffusion is non-trivial, since they operate in different feature spaces. Stable Diffusion directly conditions on the CLIP text encoder’s output, while CoDi’s audio encoder produces embeddings in the aligned CLIP space. We denote the Stable Diffusion space by \(\mathcal C_{\mathrm{SD}}\) and the aligned CLIP space by \(\mathcal C_{\mathrm{CLIP}}\). Given a text prompt \(\mathcal P\), the CLIP text encoder \(\psi_{\mathrm{text}}\) generates an SD‐space vector 
\(\mathbf c = \psi_{\mathrm{text}}(\mathcal P)\in\mathcal C_{\mathrm{SD}}\). Applying pooling to this output yields the special‐token embedding 
\(\mathbf c_{\mathrm{SD,pooled}}\in\mathcal C_{\mathrm{SD}}\). We then project and normalize:

\begin{equation}
  \label{eq:sd2codi}
  \mathbf c_{\mathrm{CLIP}}
  = \frac{\mathbf M\,\mathbf c_{\mathrm{SD,pooled}}}
         {\bigl\lVert \mathbf M\,\mathbf c_{\mathrm{SD,pooled}}\bigr\rVert},
\end{equation}
where \(\mathbf M:\mathcal C_{\mathrm{SD}}\to\mathcal C_{\mathrm{CLIP}}\) is a learned linear mapping with no bias. 

\begin{figure}[t]
    \centering
    \includegraphics[trim=2.5cm 0cm 2.5cm 0cm , width=\linewidth, clip]{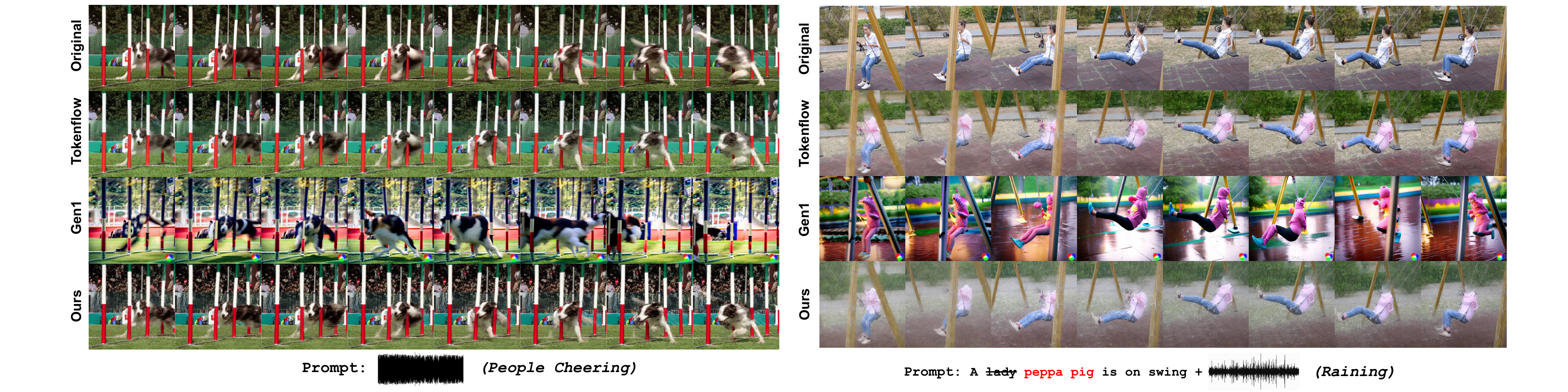}
    \caption{Editing results for Audio-guided Video Editing (\textbf{A2V-edit}) and Text and Audio-guided Video Editing (\textbf{TA2V-edit})}
    \label{fig:qualitative_video}
\end{figure}

Given an audio embedding \(\mathbf c_{A}\in\mathcal C_{\mathrm{CLIP}}\) from CoDi’s encoder \(\psi_{\mathrm{audio}}\), we invert Eq.~\ref{eq:sd2codi} and estimate its SD‐space counterpart by solving a Tikhonov‐regularized least squares problem:
\begin{equation}
  \label{eq:codi2sd}
  \tilde{\mathbf c}_{\mathrm{SD}}
  = \bigl(\mathbf M^{\!\top}\mathbf M + \lambda I\bigr)^{-1}
    \,\mathbf M^{\!\top}
    \Bigl(\lVert \mathbf c_{\mathrm{inv}}\rVert\,\mathbf c_{A}\Bigr),
\end{equation}
where \(\mathbf c_{\mathrm{inv}} = \psi_{\mathrm{text}}(\mathcal P_{\mathrm{inv}})\in\mathcal C_{\mathrm{SD}}\), \(\lambda=10^{-5}\), $P_{\mathrm{inv}}$ is the inversion prompt, and \(I\) is the identity matrix. Finally, we replicate the inverted feature $\tilde{\mathbf c}_{\text{SD}}$ certain times and concatenate it with the first and last special tokens from $\mathbf c_{\mathrm{inv}}$. Despite minor precision errors that may arise from matrix inversion, the semantic content is effectively preserved in visual editing.

\subsection{Separate Noise Branching with Adaptive Patch-wise Noise Selection}
\label{sec:Method:anytoedit:multi}
\vspace{-3pt}

Multi-modal editing commonly requires multiple guidance signals—text, audio, depth, masks, and so on—to be combined within a single diffusion process \cite{mo2024freecontrol,zhao2024uni,he2024flexecontrol,lin2024ctrl}.  
Existing works often enlarge the network or attach learned adapters, both of which incur additional training on specific paired dataset. We instead treat multi-prompt fusion as a \emph{variance-preserving aggregation} of the model’s native noise estimates, yielding a parameter-free, training-free procedure that can be plugged into any pre-trained latent diffusion model.

\vspace{1pt}
\noindent\textbf{Separate noise estimation.}
Let $\mathbf{z}_t\in\mathbb{R}^{c\times h\times w}$ be the latent at diffusion step $t$ and $\mathcal{C}=\{\mathcal{C}_1,\dots,\mathcal{C}_N\}$ the set of $N$ encoded prompts (possibly of different modalities).  
For each prompt we query the frozen denoiser
\vspace{-5pt}
\begin{equation}
    \epsilon_i=\epsilon_\theta(\mathbf{z}_t,\,t,\,\mathcal{C}_i),\qquad i=1,\dots,N.
\label{eq:snb-single}
\end{equation}

\begin{figure}[t]
    \centering
    \includegraphics[trim=0cm 0cm 0cm 0cm , width=\linewidth, clip]{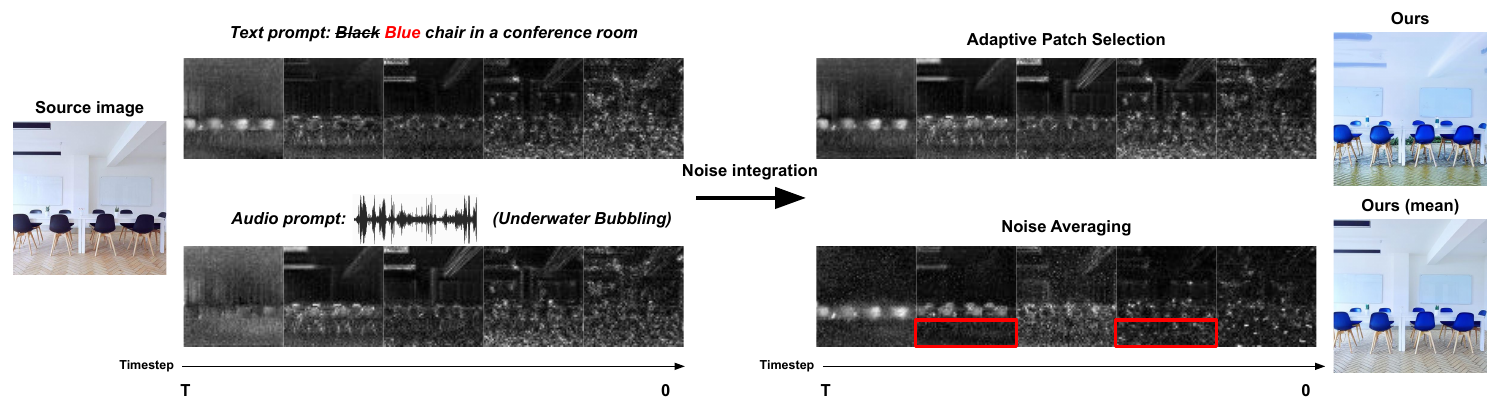}
    \caption{Comparison of different noise integration methods. Directly averaging each noise prediction diminishes important high-frequency components, as evident in the red box. In contrast, our separate noise branching and adaptive patch selection preserve these details while effectively incorporating multiple editing effects.}
    \label{fig:ablation_noise}
\end{figure}

\vspace{-3pt}
\noindent\textbf{Limitations of naïve averaging.}
A naïve fusion, $\hat\epsilon_\theta=\tfrac1N\sum_i\epsilon_i$, reduces variance in proportion to $1/N$.  
Because diffusion process relies on scaled noise to reconstruct detail, this variance collapse produces suboptimal results, as observed in Fig.~\ref{fig:ablation_noise} (top) and quantified in Sec.~\ref{sec:Experiments}.

\vspace{1pt}
\noindent\textbf{Adaptive patch-wise Noise Selection.}
We utilize the inversion prompt $\mathcal{C}_\text{inv}$ and its noise prediction $\epsilon_\text{inv}=\epsilon_\theta(\mathbf{z}_t,\,t,\,\mathcal{C}_\text{inv})$ to migitage this limitation. Subtracting $\epsilon_\text{inv}$ from each editing prompt yields residual feature maps $\Delta\epsilon_i=\epsilon_i-\epsilon_\text{inv}$, where each $ \Delta \epsilon_i \in \mathbb{R}^{c \times h \times w} $ represents the difference in noise predictions for each prompt. Let $p$ index spatial locations on the $h\times w$ grid. We retain, for every patch, the residual with the largest magnitude:
\begin{equation}
\hat\epsilon_\theta(p)=\epsilon_\text{inv}(p)+\Delta\epsilon_{i^\star(p)}(p),\quad
i^\star(p)=\arg\max_i\lVert\Delta\epsilon_i(p)\rVert
\label{eq:snb-max}
\vspace{-10pt}
\end{equation}
, where $\lVert\cdot\rVert$ is the channel-wise $\ell_2$ norm.  
Equation~\eqref{eq:snb-max} preserves high-frequency detail and prevents destructive interference between heterogeneous conditions, as illustrated in Fig.~\ref{fig:ablation_noise} (bottom).

\section{Experiments} \label{sec:Experiments}
In this section, we rigorously validate the effectiveness of our framework by constructing various evaluation datasets and baselines that incorporate audio editing prompts. Specifically, quantitative evaluations are conducted across four tasks: Audio-guided Image Editing (\textbf{A2I-edit}), Text and Audio-guided Image Editing (\textbf{TA2I-edit}), Audio-guided Video Editing (\textbf{A2V-edit}), and Text and Audio-guided Video Editing (\textbf{TA2V-edit}). Additionally, we explore more complex editing scenarios in Section \ref{sec:Experiments:analysis}.

The results from the A2I-edit and A2V-edit tasks assess whether our audio mapping can produce appropriate editing results while preserving the diverse information contained in the audio. The TA2I-edit and TA2V-edit tasks demonstrate that our noise integration method yields suitable results in complex editing scenarios that existing visual editing approaches cannot handle. The zero-shot capability of our framework can be further verified in the Sections 3, 4, 5, and 6 of the supplementary material.

\subsection{Implementation Details} \label{sec:Experiments:implementation}
We incorporate diffusion editing techniques such as PnP-Diffusion \cite{tumanyan2023plug} and TokenFlow \cite{geyer2023tokenflow}. These techniques demonstrate superior edit fidelity and do not require word-level configurations, such as specifying which words to replace or emphasize. We employ Stable Diffusion v1.5 for fairness and utilize A100 with 40GB memory for inference with the seed set to 1 for all experiments. 

\subsection{Evaluation Dataset} \label{sec:Experiments:dataset}
To perform a comprehensive evaluation of our framework, we introduce \emph{PIEBench-multi} and \emph{DAVIS-multi}, which augment existing text-guided visual editing benchmarks, PIEBench \cite{ju2023direct} and DAVIS \cite{pont20172017} to involve audio editing prompts. We select 28 classes of audio suitable for editing by filtering out low-quality content from VGGSound \cite{chen2020vggsound} and choose three audio clips for each category. We utilize 100 images in the PIEBench dataset and use all 89 videos from DAVIS, excluding one unsuitable video (gunshot). We then match appropriate audio editing prompts to each visual data, resulting in 300 editing pairs in PIEBench-multi and 267 editing pairs in DAVIS-multi. Detailed examples of our editing pairs used in the benchmarks can be found in Section 1 of the supplementary material.

\subsection{Baselines} \label{sec:Experiments:baselines}
Complex audio-guided visual editing is underexplored and lacks baselines, so we first compare our method to representative text-guided approaches: Prompt-to-Prompt (P2P) \cite{hertz2022prompt}, Null-Text Inversion (NT) \cite{mokady2023null}, and PnP-Diffusion (PnP) \cite{tumanyan2023plug} for image editing; and TokenFlow \cite{geyer2023tokenflow} and Gen-1 \cite{esser2023structure} for video editing.

We initially consider a straightforward baselines that uses audio class names either alone (A2I-edit, A2V-edit) or appended to text editing prompts (TA2I-edit, TA2V-edit). Since this approach does not incorporate the textual description \(\mathcal{P}_{\text{inv}}\) of original data, it induces severe modifications and produces suboptimal results. Section~2 of the supplementary material presents complete performances and examples for these baselines.

To better integrate audio information, we employ an audio captioning model \cite{kim2024enclap} to convert audio into text. Then we refine \(\mathcal{P}_{\text{inv}}\) with audio caption using GPT-4o \cite{achiam2023gpt}. Section~2 of the supplementary material provides details on constructing these refined baselines, which we denote as X(caption).

For A2I-edit task, we compare our approach with AVStyle \cite{li2022learning}, Sonic Diffusion \cite{biner2024sonicdiffusion}, and caption-based baselines. We exclude SGSIM \cite{lee2022sound} as it is trained with highly specific domains and tends to produce inappropriate results on our benchmark. For TA2I-edit we include caption-based baselines, Sonic Diffusion, and 12 cascaded pipelines that sequentially run a T2I-edit (P2P, NT, or PnP) and then an A2I-edit (AVStyle or Sonic Diffusion). For A2V-edit and TA2V-edit tasks, we select baselines only using caption-based approaches (TokenFlow and Gen-1) since there are no suitable audio-guided video editing methods on our benchmarks.

\begin{table*}[t]
    \centering
    \captionsetup{width=\linewidth}
    \resizebox{\linewidth}{!}{%
    \begin{tabular}{llccccc|cccccccccccc}
        \toprule
        && \textbf{Str. Dist$(\downarrow)$} & \textbf{LPIPS$(\downarrow)$} & \textbf{CLIP\_Audio$(\uparrow)$} & \textbf{CLIP\_Text$(\uparrow)$} & \textbf{CLIP\_integrated$(\uparrow)$} & \multicolumn{3}{c}{\textbf{Structural Preservation$(\uparrow)$}} & \multicolumn{3}{c}{\textbf{Audio Alignment$(\uparrow)$}} & \multicolumn{3}{c}{\textbf{Text Alignment$(\uparrow)$}} & \multicolumn{3}{c}{\textbf{Overall Alignment$(\uparrow)$}} \\
        && & & & & & Win & Tie & Lose & Win & Tie & Lose & Win & Tie & Lose & Win & Tie & Lose \\
        \midrule
        \multirow{6}{*}{\textbf{A2I-edit}} 
            & AVStyle & 0.0396 & 0.2219 & 14.5596 & - & - & \textbf{88.40} & 7.60 & 4.00 & \textbf{90.06} & 2.59 & 7.34 & - & - & - & - & - & -  \\
            & Sonic Diffusion & 0.0475 & 0.2915 & 13.4908 & - & - & \textbf{77.38} & 4.96 & 17.66 & \textbf{91.27} & 4.15 & 4.59 & - & - & - & - & - & - \\
            & P2P (caption) & 0.0617 & 0.3261 & 12.2536 & - & - & \textbf{89.66} & 2.87 & 7.47 & \textbf{82.31} & 11.13 & 6.56 & - & - & - & - & - & - \\
            & NT (caption) & \textbf{0.0097} & \textbf{0.1145} & 12.7248 & - & - & \textbf{78.85} & 10.04 & 11.11 & \textbf{86.41} & 8.70 & 4.88 & - & - & - & - & - & - \\
            & PnP (caption) & 0.0227 & 0.1866 & 12.8969 & - & - & \textbf{74.02} & 3.52 & 22.46 & \textbf{75.04} & 8.56 & 16.40 & - & - & - & - & - & - \\
            & \textbf{Ours} & 0.0341 & 0.2539 & \textbf{17.0472} & - & - & - & - & - & - & - & - & - & - & - & - & - & - \\
        \midrule
        \multirow{7}{*}{\textbf{TA2I-edit}} 
            & AVStyle $\rightarrow$ NT & 0.0668 & 0.4174 & 13.1312 & 25.2388 & 21.1613 & \textbf{68.69} & 5.05 & 26.26 & \textbf{70.42} & 5.52 & 24.06 & \textbf{58.92} & 11.42 & 29.66 & \textbf{70.88} & 13.65 & 15.48  \\
            & NT $\rightarrow$ AVStyle & 0.0369 & 0.2508 & 14.4867 & \textbf{25.5348} & 21.6772 & \textbf{56.59} & 14.10 & 29.31 & \textbf{71.73} & 20.90 & 7.36 & \textbf{46.12} & 14.88 & 38.99 & \textbf{59.15} & 17.41 & 23.44  \\
            & P2P (caption) & 0.0591 & 0.3278 & 12.3059 & 24.9569 & 21.1932 & \textbf{56.47} & 8.04 & 35.49 & \textbf{69.48} & 11.69 & 18.83 & \textbf{48.64} & 14.40 & 36.96 & \textbf{76.60} & 7.35 & 16.05  \\
            & NT (caption) & 0.0192 & 0.1802 & 12.8379 & 24.3057 & 21.6057 & \textbf{56.59} & 14.10 & 29.31 & \textbf{71.73} & 20.90 & 7.36 & \textbf{46.12} & 14.88 & 38.99 & \textbf{59.15} & 17.41 & 23.44  \\
            & PnP (caption) & 0.0264 & 0.2108 & 13.0695 & 24.9375 & 21.3103 & \textbf{50.83} & 16.39 & 32.78 & \textbf{76.88} & 7.89 & 15.23 & \textbf{48.99} & 17.31 & 33.70 & \textbf{48.12} & 27.33 & 24.55  \\
            & Sonic Diffusion & \textbf{0.0179} & \textbf{0.1551} & 12.9011 & 23.6654 & 20.1574 & \textbf{49.80} & 9.92 & 40.28 & \textbf{63.12} & 14.17 & 22.71 & \textbf{51.22} & 17.64 & 31.14 & \textbf{69.61} & 12.53 & 17.87  \\
            & \textbf{Ours (mean)} & 0.0204 & 0.1729 & 14.0025 & 24.2257 & 20.1574 & 39.69 & 18.51 & \textbf{41.79} & \textbf{55.33} & 22.79 & 21.88 & \textbf{27.36} & 48.69 & 23.94 & \textbf{39.71} & 39.92 & 20.37  \\
            & \textbf{Ours} & 0.0333 & 0.2392 & \textbf{15.0699} & 24.1283 & \textbf{23.0356} & - & - & - & - & - & - & - & - & - & - & - & -\\
        \bottomrule
    \end{tabular}
    }
    \caption{
        Evaluation results for \textbf{A2I-edit} and \textbf{TA2I-edit}. Left: quantitative scores; right: user study.
        \textbf{Str. Dist} is structure distance.
        \textbf{CLIP\_Audio}, \textbf{CLIP\_Text}, and \textbf{CLIP\_integrated} are CLIP scores w.r.t. audio, text, and integrated captions.
        User ratings—Structural Pres., Audio Align., Text Align., Overall—compare ours with the baseline; bold indicates the best result.
    }
    \label{tab:image_edit_results}
\end{table*}

\subsection{Evaluation Metrics} \label{sec:Experiments:metrics}
To evaluate our framework, we employ metrics for content preservation and edit fidelity following existing works \cite{hertz2022prompt, mokady2023null, tumanyan2023plug, geyer2023tokenflow}. For content preservation, we use Structure Distance \cite{tumanyan2022splicing} and LPIPS \cite{zhang2018unreasonable} for image editing tasks on PIEBench-multi (A2I-edit and TA2I-edit). For video editing tasks on DAVIS-multi (A2V-edit and TA2V-edit), we utilize frame-wise CLIP similarity (CLIP\_frame) in accordance with the TGVE competition \cite{wu2023cvpr}.

To evaluate edit fidelity, we apply CLIPScore \cite{hessel2021clipscore} across all tasks. Specifically, we calculate CLIPScore for each individual editing prompt (CLIP\_audio and CLIP\_text) and use category names for audio editing prompts. However, in editing tasks with multiple prompts, it is crucial to evaluate how well each prompt is integrated and applied together. To evaluate this, we manually create integrated captions in text form for these tasks and report the results as CLIP\_integrated.

Although these metrics assess the preservation of original data and the effectiveness of editing effects, they do not provide a comprehensive assessment. To address this limitation, we conducted a user study on Amazon Mechanical Turk with five participants per question, using a total of 100 samples for image editing tasks and all available samples for video editing tasks. Each participant compared our method against a baseline, selecting the better model according to structural preservation, audio alignment, text alignment, and overall condition consistency. To prevent cases where the absence of edits was rated highly for structural preservation, participants were instructed to give the lowest score for structural preservation if no edits were observed.

\begin{table*}[t]
    \centering
    \captionsetup{width=\linewidth}
    \resizebox{\linewidth}{!}{%
    \begin{tabular}{llcccc| ccc ccc ccc ccc}
        \toprule
        & & \textbf{CLIP\_frame$(\uparrow)$} & \textbf{CLIP\_Audio$(\uparrow)$} & \textbf{CLIP\_Text$(\uparrow)$} & \textbf{CLIP\_integrated$(\uparrow)$} & \multicolumn{3}{c}{\textbf{Structural Preservation$(\uparrow)$}} & \multicolumn{3}{c}{\textbf{Audio Alignment$(\uparrow)$}} & \multicolumn{3}{c}{\textbf{Text Alignment$(\uparrow)$}} & \multicolumn{3}{c}{\textbf{Overall Alignment$(\uparrow)$}} \\
        && & & & & Win & Tie & Lose & Win & Tie & Lose & Win & Tie & Lose & Win & Tie & Lose \\
        \midrule
        \multirow{3}{*}{\textbf{A2V-edit}} 
            & Tokenflow (caption) & \textbf{0.9060} & 15.2095 & - & - & \textbf{61.98} & 8.89 & 29.14 & \textbf{61.25} & 10.47 & 28.29 & - & - & - & - & - & - \\
            & Gen1 (caption) & 0.8854 & 14.5752 & - & - & \textbf{57.07} & 2.31 & 40.62 & \textbf{70.77} & 6.52 & 22.71 & - & - & - & - & - & - \\
            & \textbf{Ours} & 0.9029 & \textbf{18.1445} & - & - & - & - & - & - & - & - & - & - & - & - & - & - \\
        \midrule
        \multirow{4}{*}{\textbf{TA2V-edit}} 
            & Tokenflow (caption) & \textbf{0.9162} & 13.2351 & 23.6890 & 22.4227 & \textbf{43.50} & 19.28 & 37.22 & \textbf{54.84} & 9.30 & 35.86 & \textbf{49.78} & 21.34 & 28.88 & \textbf{58.40} & 9.02 & 32.58  \\
            & Gen1 (caption) & 0.8780 & 14.8816 & \textbf{23.7246} & 22.8530 & \textbf{54.46} & 0.48 & 45.06 & \textbf{48.41} & 10.72 & 40.87 & \textbf{50.39} & 5.17 & 44.44 & \textbf{49.79} & 8.58 & 41.63  \\
            & \textbf{Ours (mean)} & 0.9082 & 15.4571 & 22.9909 & 22.7029 & \textbf{48.67} & 10.40 & 40.93 & \textbf{50.54} & 37.91 & 11.55 & \textbf{35.46} & 39.38 & 25.15 & \textbf{38.19} & 38.56 & 23.25  \\
            & \textbf{Ours} & 0.9043 & \textbf{16.1833} & 22.8336 & \textbf{23.2885} & - & - & - & - & - & - & - & - & - & - & - & -  \\
        \bottomrule
    \end{tabular}
    }
    \caption{
        Evaluation results for Audio-guided Video Editing (A2V-edit) and combined Text and Audio-guided Video Editing (TA2V-edit). 
        \textbf{CLIP\_frame} represents the frame-wise CLIP similarity score. 
        All other abbreviations are consistent with those used in Table \ref{tab:image_edit_results}.
    }
    \label{tab:video_edit_results}
\end{table*}

\subsection{Evaluation Results} \label{sec:Experiments_results}
Table \ref{tab:image_edit_results} shows the quantitative evaluation results of A2I-edit and TA2I-edit on PIEBench-multi, and Table \ref{tab:video_edit_results} presents the quantitative evaluation results of A2V-edit and TA2V-edit on DAVIS-multi. We also show qualitative comparisons in Figure~\ref{fig:qualitative_image} and \ref{fig:qualitative_video}.

\textbf{Audio-guided editing.} Our model significantly outperforms the baselines in A2I-edit and A2V-edit in terms of CLIP\_audio. Our method does not show competitive scores in content preservation metrics (Structure Distance, LPIPS, and CLIP\_frame) since the baselines tend to preserve the original data excessively and do not well reflect the editing prompt. 

Qualitative samples in Figure \ref{fig:qualitative_image} and Figure \ref{fig:qualitative_video} allow us to observe these characteristics in detail. In Fig.~\ref{fig:qualitative_image} (top), our method converts a bird to an airplane and adds a volcanic blast to a mountain, while baselines fail. In Fig.~\ref{fig:qualitative_video} (left), our model inserts the crowd following the audio prompt, whereas baselines scarcely change. More samples of the zero-shot capabilities of our framework can be found in Section 3 and Section 5 of the supplementary material. These results demonstrate that our framework effectively incorporates the rich information from audio and achieves diverse editing effects that are difficult to capture using text alone. 

\textbf{Audio and Text-guided editing.} Similarly, our model outperforms the baselines in the CLIP\_audio and CLIP\_integrated. Baselines score well in CLIP\_text but drop sharply in other metrics, showing that they capture only part of the desired edits in complex cases, a phenomenon also reported by \cite{kumari2023multi, brack2024ledits++}.

In Figure \ref{fig:qualitative_image} (row 3), our method converts the dog into a rabbit and adds the splashing-water effect that baselines miss. In row 4, it morphs the woman into a blue-made-up baby, while baselines merely recolor the makeup. In Figure \ref{fig:qualitative_video} (right) our model replaces the lady with Peppa Pig under rainfall, while baselines either omit the rain or over-distort the scene. Additional examples appear in Section 4 and Section 6 of the supplementary material, underscoring how separate-noise branching and adaptive patch selection effectively applies editing effects in complex scenarios.

\textbf{User Study.} Tables \ref{tab:image_edit_results} and \ref{tab:video_edit_results} present the results of the user study, demonstrating that our approach significantly outperforms the baselines in all cases, except for slightly lower performance than averaged noise prediction in terms of structural preservation. Although many baselines fail to produce any editing effects in several cases, users were instructed to account for such overly preserved images with the guideline: \textit{EXCEPTION: if an image is unchanged from the original image, consider it as the worst case.} The results demonstrate that our method effectively edits images and videos in alignment with both text and audio while maintaining the original structure.


Note that in the main paper, only AVStyle $\rightarrow$ NT and NT $\rightarrow$ AVStyle were included among the 12 cascaded approach baselines for TA2I-edit, as these two methods demonstrate strong performance in A2I-edit and also due to space. We include the results of all baselines in Section 7 of the supplementary material. Also, we observe that CLIP\_text reports higher scores than CLIP\_audio since the text description of input data is included in the text editing prompt.

\begin{figure}[t]
\centering            
\begin{tabular}{cc}   
  (a) & (b)
  \\
  \bmvaHangBox{%
    \includegraphics[width=0.55\linewidth]{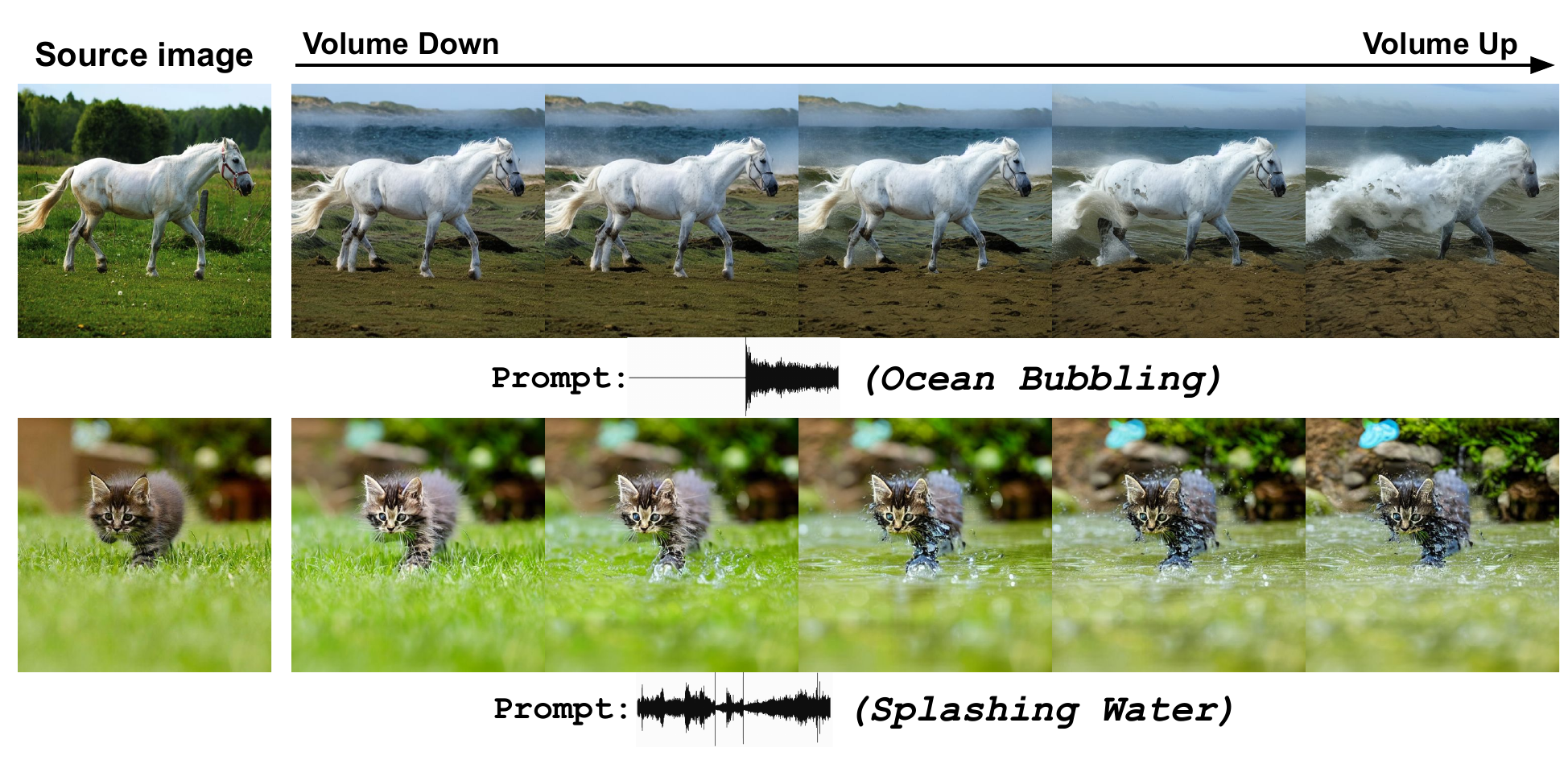}}
  &
  \bmvaHangBox{%
    \includegraphics[width=0.36\linewidth]{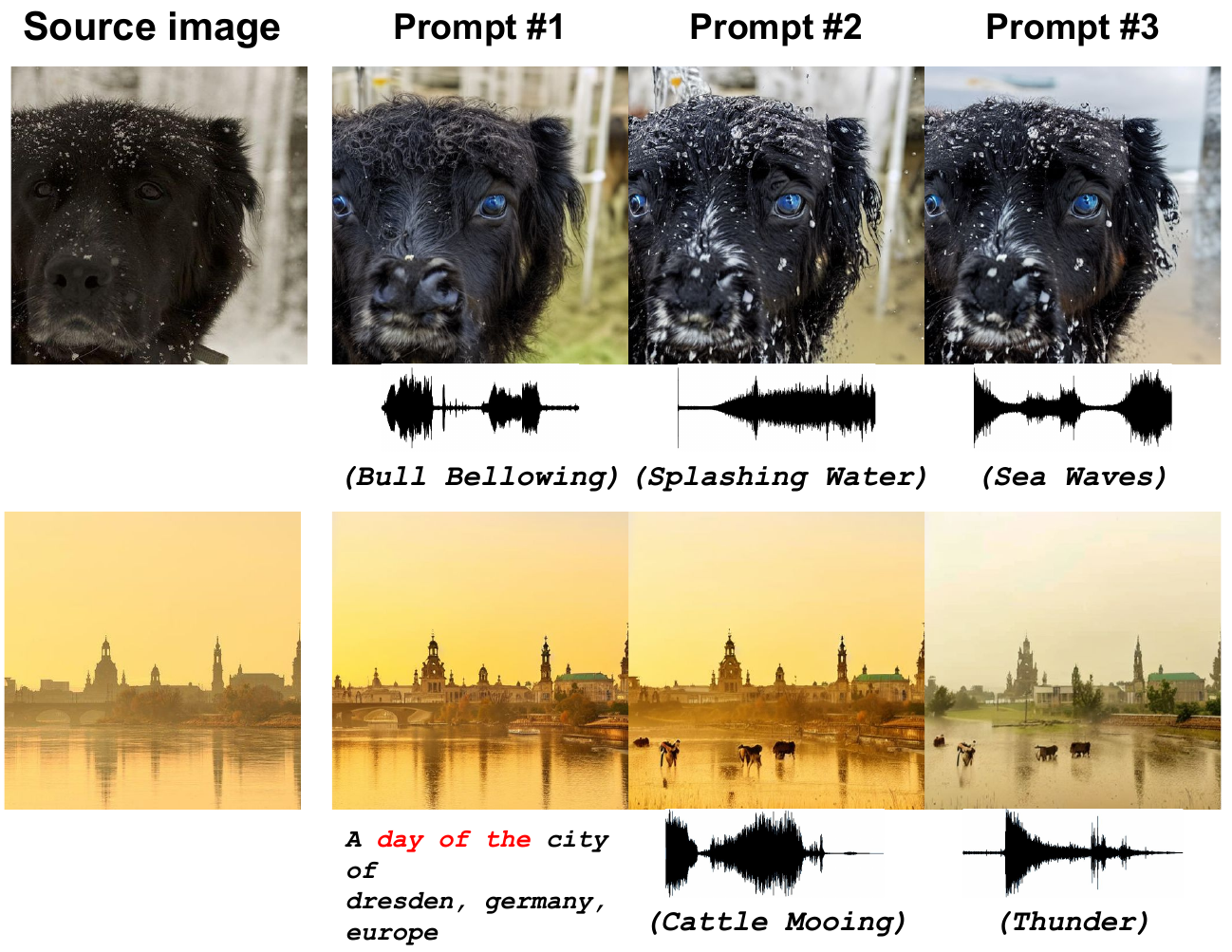}}
\end{tabular}

\caption{Audio-guided visual editing examples with complex settings—(a) magnitude guided effects and (b) multi-prompt editing.}
\label{fig:two_cols_bmvc}
\end{figure}

\subsection{Ablation Study} \label{sec:Experiments:ablation}
To demonstrate the effectiveness of our noise integration approach, we conduct an ablation study. Specifically, we compare our method to one that integrates multiple noise predictions by simply averaging them. We report these results as `Ours (mean)' in Tables \ref{tab:image_edit_results} and Table \ref{tab:video_edit_results}. Simply averaging (mean) each noise prediction results in a drastic performance drop in CLIP\_integrated scores across all tasks, which indicates that our separate noise branching and adaptive patch selection play a crucial role in handling complex multiple editing prompts. 

Furthermore, Figure \ref{fig:ablation_noise} visualizes the differences between the two methods when integrating multiple editing prompts. Specifically, we plot the magnitudes of the separate noise predictions from each editing prompt and the integrated noise at each diffusion timestep. As indicated by the red boxes, simply averaging the noise predictions causes the information from the audio editing prompt to be canceled out, resulting in the \textit{underwater bubbling} effect not appearing in the final output.

\subsection{Analysis and Discussions} \label{sec:Experiments:analysis}
Figure \ref{fig:two_cols_bmvc}(a) illustrates the impact of modulating the magnitude of the audio, showing that an increase in magnitude enhances the editing effect. This demonstrates that our framework effectively integrates rich audio information such as magnitude, which is difficult to capture using text alone.

Figure \ref{fig:two_cols_bmvc}(b) illustrates examples of more complex visual editing scenarios involving more than two editing prompts. We add text or audio editing prompts one at a time and visualize each result. We observe that our framework effectively determines which prompts should influence the objects and which should affect the background. Additional examples are provided in Section 8 of our supplementary material.

As for limitations, our frameworks inherit certain constraints from PnP-Diffusion and TokenFlow, particularly a tendency to preserve composition and content excessively or to yield flickering outputs. Although we did not address this in the present study, combining our framework with more powerful diffusion models and editing methods would also be worthwhile. Additionally, our mapping function assumes that audio features align closely with textual counterparts, which may not always be accurate \cite{liang2022mind}. Utilizing a variety of multi-modal encoders and diffusion models also appears to be a promising direction for future research.
\section{Conclusion}
\label{sec:conclusion}
\vspace{-5pt}
In this paper, we introduce a framework for visual editing that seamlessly integrates audio conditions into diffusion models without requiring additional training. By leveraging a pre-trained multi-modal encoder with strong zero-shot capabilities, our approach mitigates the limitations of editing methods with only textual guidance, especially in complex scenarios where textual descriptions would be insufficient or ambiguous. We further introduce separate noise branching and adaptive patch selection, a novel method to handle multiple and multi-modal editing prompts without further training. Extensive experiments on newly established benchmark datasets demonstrate that our framework surpasses existing methods, effectively capturing rich audio information and coherently combining multiple prompts to produce sophisticated and accurate visual editing.

\section{Acknowledgement}
This work was supported by Institute of Information \& communications Technology Planning \& Evaluation (IITP) grant funded by the Korea government (MSIT) (No.2022-0-00608, Artificial intelligence research about multi-modal interactions for empathetic conversations with humans \& No.RS-2020- II201336, Artificial Intelligence graduate school support(UNIST)). This work was also supported by National IT Industry Promotion Agency (NIPA) grant funded by the Korea government (MSIT) (No. PJT-25-033312, Development and Demonstration of a Fully Autonomous Speed Sprayer (SS) Based on On-Device AI).

\bibliography{reference}
\end{document}


\setcounter{page}{1}
In the supplementary material, we provide additional samples and experimental results that we were unable to include fully due to space constraints. All samples can be examined in more detail via the attached web page (index.html).

\section{Samples from PIEBench-multi and DAVIS-multi}\label{sec:sup_benchmark}

\begin{figure}[!htbp]
    \centering
    \captionsetup{width=\textwidth}
    \includegraphics[page=1, width=0.85\linewidth]{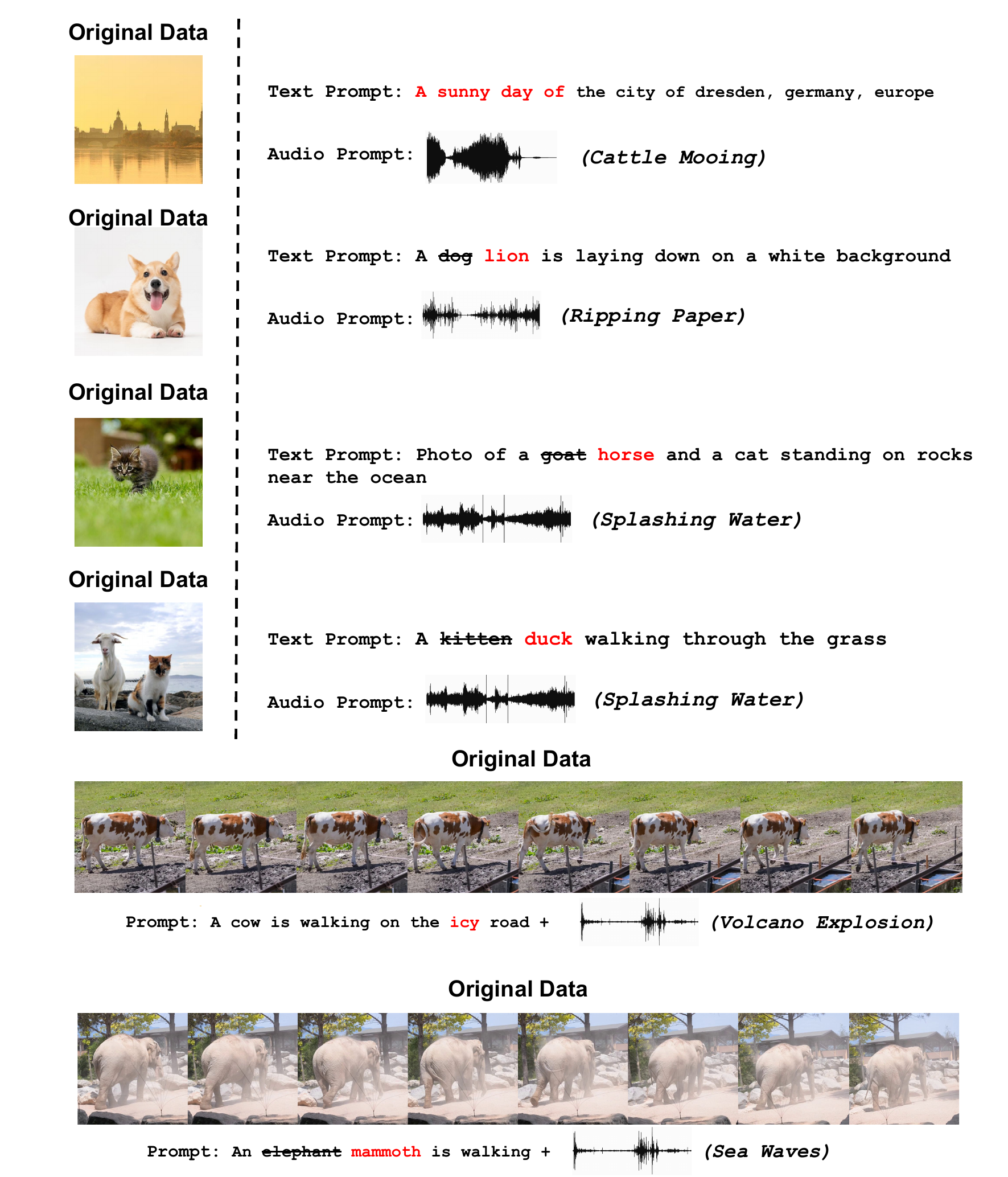}
    \caption{Samples from PIEBench-multi and DAVIS-multi}
\end{figure}
\FloatBarrier

\newpage
\section{Detailed Description of Baselines Construction}\label{sec:sup_benchmark}

\begin{table}[ht]
    \centering
    \captionsetup{width=\linewidth}
    \vspace{-1.4em}
    \resizebox{0.6\linewidth}{!}{%
        \begin{tabular}{llcccccc}
            \toprule
            && \textbf{Str. Dist$(\downarrow)$} & \textbf{LPIPS$(\downarrow)$} & \textbf{CLIP\_Audio$(\uparrow)$} & \textbf{CLIP\_Text$(\uparrow)$} & \textbf{CLIP\_integrated$(\uparrow)$} \\
            \midrule
            \multirow{3}{*}{\textbf{A2I-edit}} 
                & NT (keyword) & 0.0407 & 0.2993 & 17.0035 & - & - \\
                & \textbf{Ours} & \textbf{0.0341} & \textbf{0.2539} & \textbf{17.0472} & - & - \\
            \midrule
            \multirow{3}{*}{\textbf{TA2I-edit}} 
                & NT (keyword) & 0.0934 & 0.4530 & \textbf{15.7799} & 22.6774 & 22.8283 \\
                & \textbf{Ours} & \textbf{0.0333} & \textbf{0.2392} & 15.0699 & \textbf{24.1283} & \textbf{23.0356} \\
            \bottomrule
        \end{tabular}
    }
    \caption{
        Image editing performance using audio class names.
    }\label{tab:image_edit_results_rebuttal}
\end{table}

\begin{table}[ht]
    \centering
    \captionsetup{width=\linewidth}
    \resizebox{0.6\linewidth}{!}{%
        \begin{tabular}{llccccccccc}
            \toprule
            & & \textbf{LPIPS$(\downarrow)$} & \textbf{CLIP\_Audio$(\uparrow)$} & \textbf{CLIP\_Text$(\uparrow)$} & \textbf{CLIP\_integrated$(\uparrow)$} \\
            \midrule
            \multirow{3}{*}{\textbf{A2V-edit}} 
                & Tokenflow (keyword) & 0.3879 & 17.8147 & - & - \\
                & \textbf{Ours} & \textbf{0.2317} & \textbf{18.1445} & - & - \\
            \midrule
            \multirow{3}{*}{\textbf{TA2V-edit}} 
                & Tokenflow (keyword) & 0.3476 & 13.6984 & \textbf{23.5546} & 22.6514 \\
                & \textbf{Ours} & \textbf{0.2064} & \textbf{16.1833} & 22.8336 & \textbf{23.2885} \\
            \bottomrule
        \end{tabular}
    }
    \caption{
        Video editing performance using audio class names.
    }\label{tab:video_edit_results_rebuttal}

\end{table}

\begin{figure}[!htbp]
    \centering
    \captionsetup{width=\textwidth}
    \includegraphics[page=1, width=\linewidth]{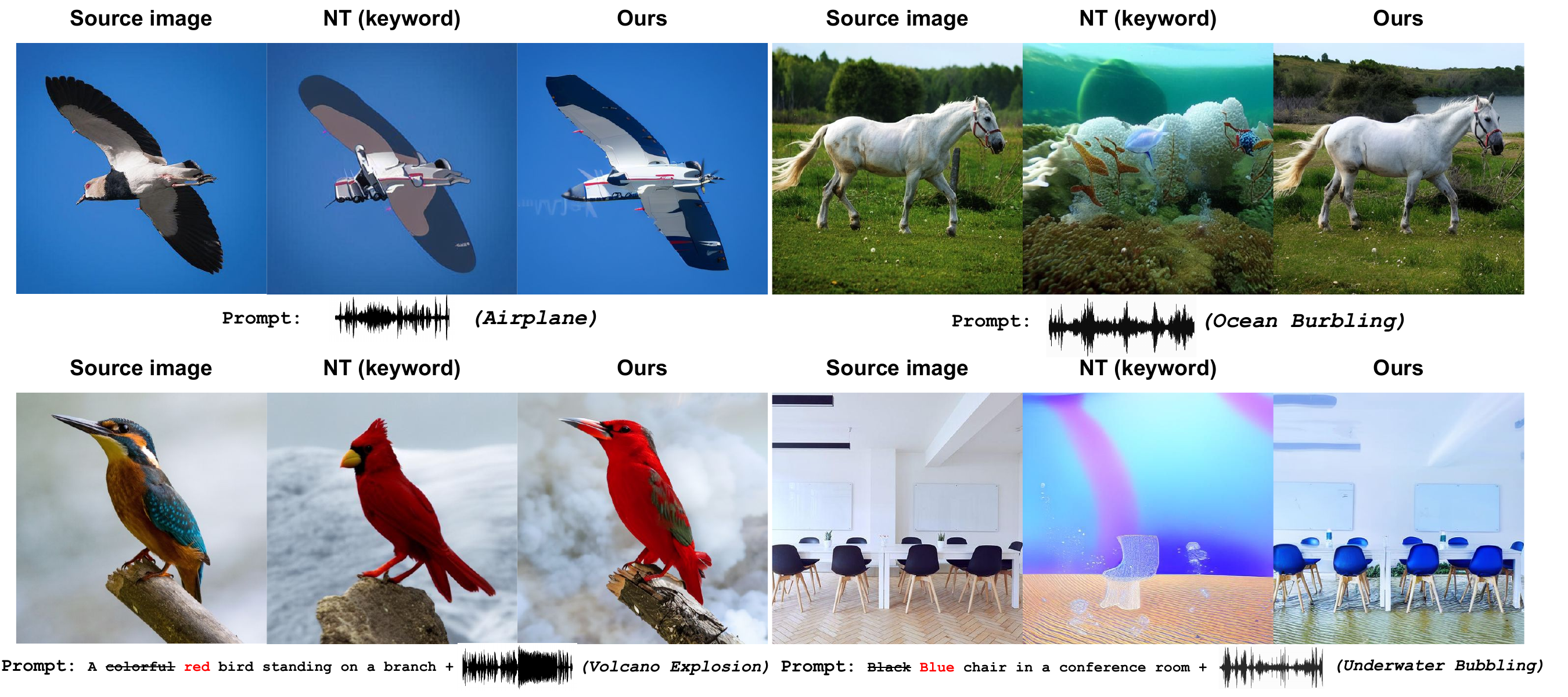}
    \caption{Qualitative comparison of our model and NT(keyword)}
    \label{fig:keyword_supp}
\end{figure}

Due to the absence of comparable models for our approach, we constructed a baseline using well-established text-guided editing methods. Specifically, we used VGGSound audio class names as editing prompts (A2I-edit, A2V-edit) or appended them to text prompts (TA2I-edit, TA2V-edit). We refer to these collectively as “X (keyword).” As shown in Tables~\ref{tab:image_edit_results_rebuttal} and \ref{tab:video_edit_results_rebuttal} and Figure~\ref{fig:keyword_supp}, these approaches altered the original data too drastically, making them unsuitable as a proper baseline.

To establish a more competitive baseline for visual editing that effectively leverages audio information, we constructed a more sophisticated approach that combines textual descriptions of the input data with audio. First, we employ EnCLAP to convert the audio into text. Then, we used GPT-4o to merge the two textual inputs, with the following specific prompt:

\textit{"You will be given two texts: one is a base text that describes a scene, while another describes the audio in the scene. Your job is to add audio description naturally to the scene description. The scene description itself must not be changed. Only output the integrated sentence."}.

\clearpage
\section{Further Samples on A2I-edit}\label{sec:sup_a2i}

\begin{figure}[!htbp]
    \centering
    \captionsetup{width=\textwidth}
    \includegraphics[page=1, width=\linewidth]{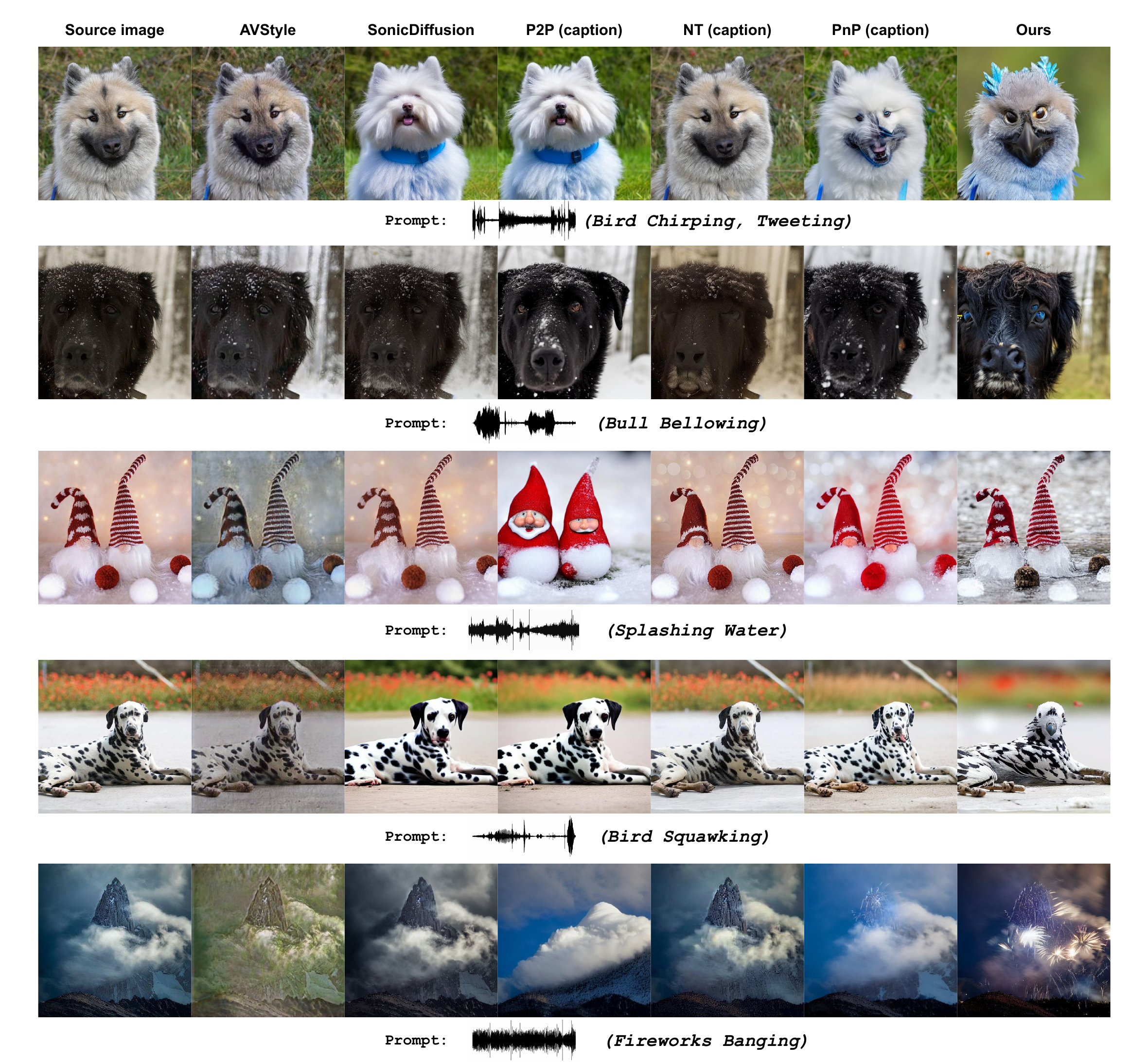}
    \caption{Samples from A2I-edit - Page 1}
\end{figure}
\FloatBarrier

\begin{figure*}[!htbp]
    \centering
    \captionsetup{width=\textwidth}
    \includegraphics[page=2, width=\linewidth]{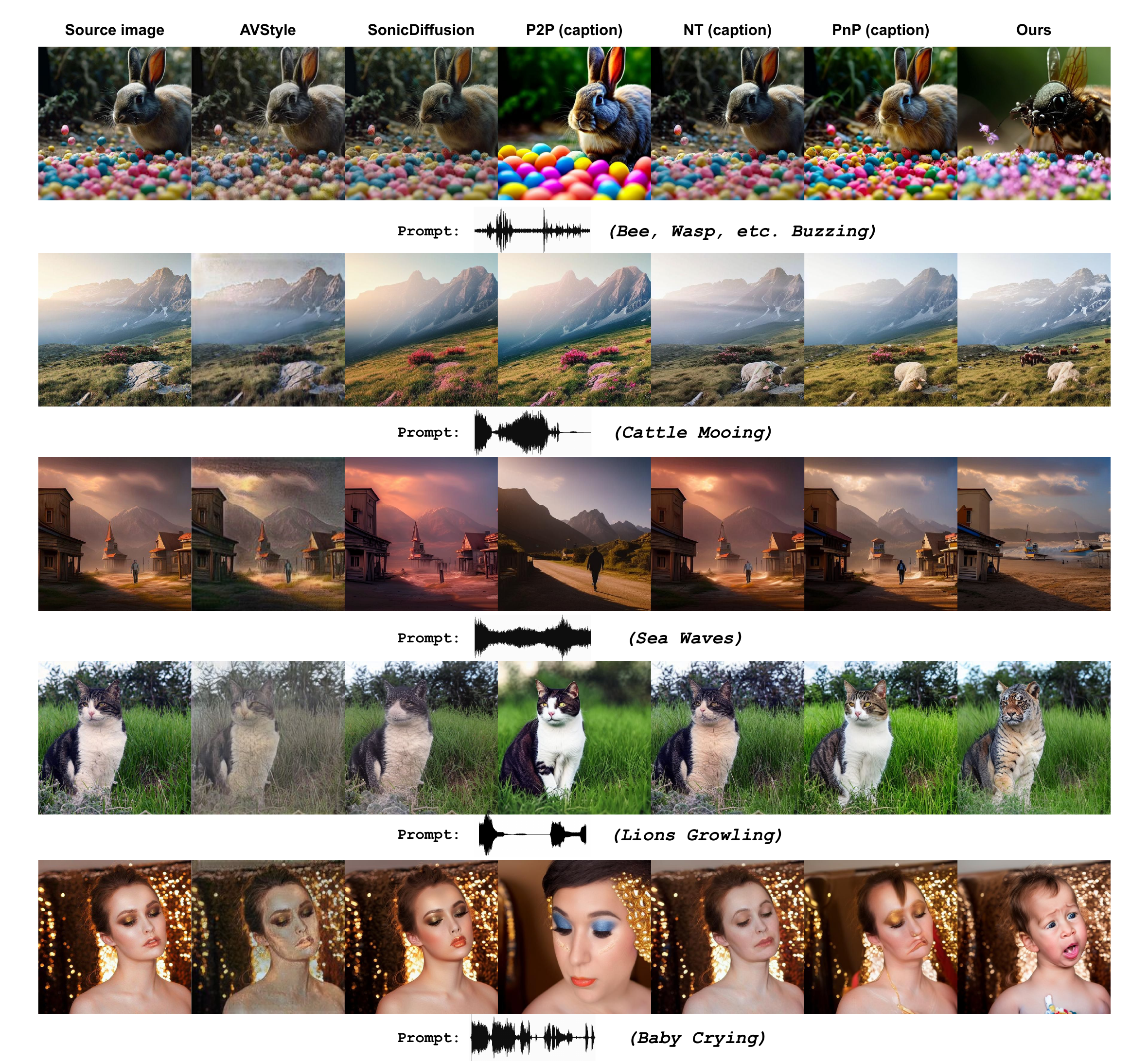}
    \caption{Samples from A2I-edit - Page 2}
\end{figure*}

\begin{figure*}[!htbp]
    \centering
    \captionsetup{width=\textwidth}
    \includegraphics[page=3, width=\linewidth]{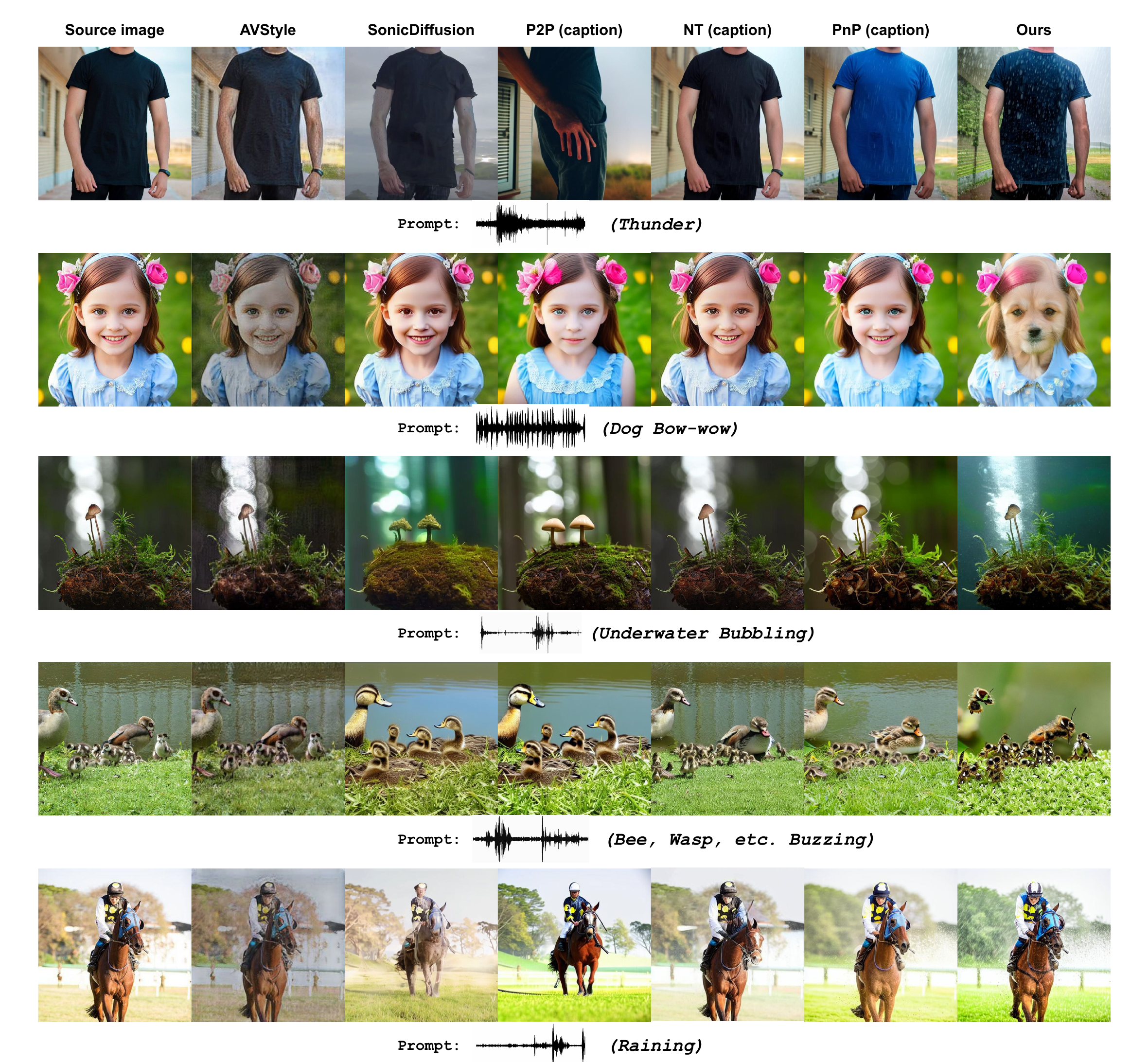}
    \caption{Samples from A2I-edit - Page 3}
\end{figure*}

\clearpage
\section{Further Samples on TA2I-edit}\label{sec:sup_ta2i}

\begin{figure*}[!htbp]
    \centering
    \captionsetup{width=\textwidth}
    \includegraphics[page=1, width=\linewidth]{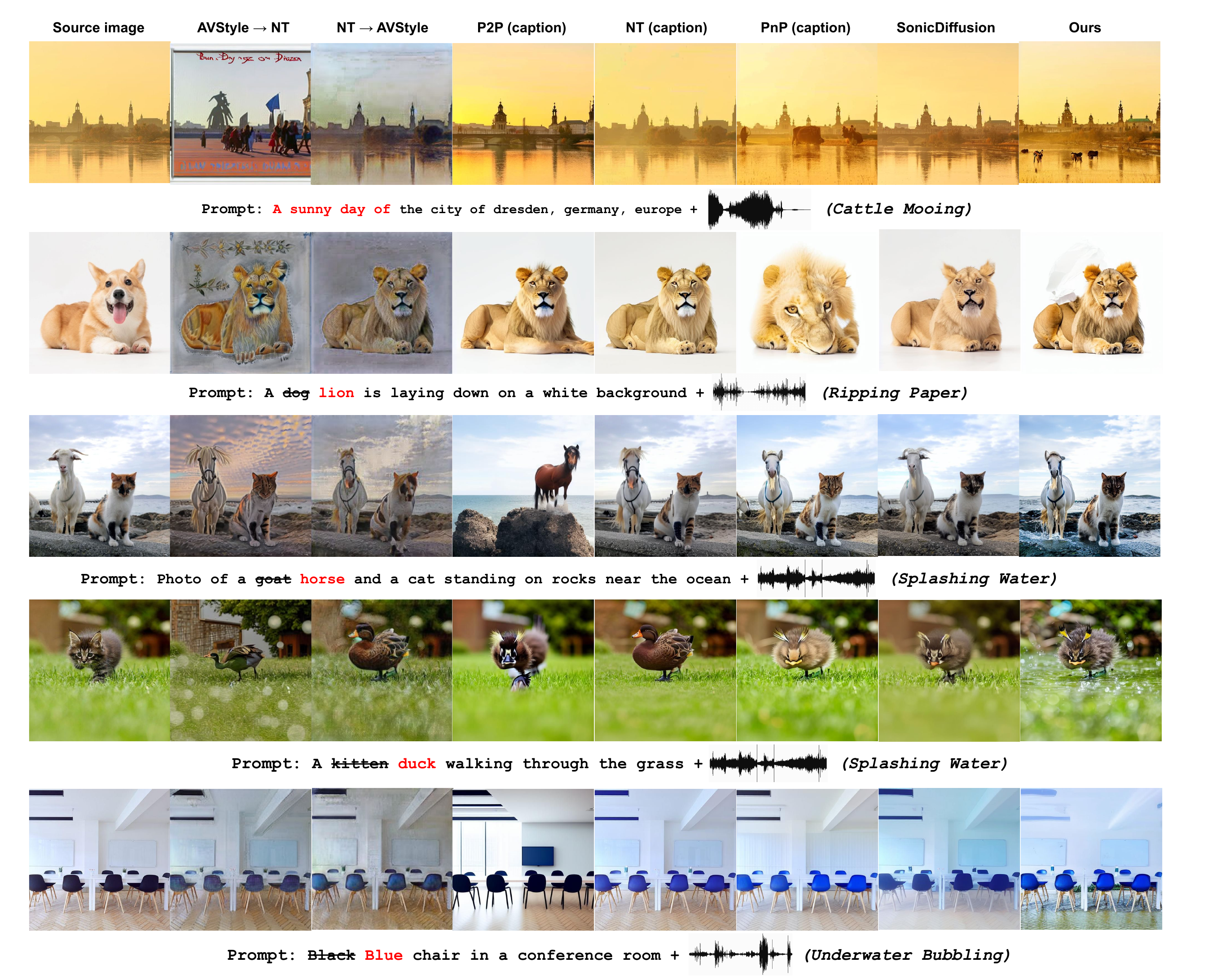}
    \caption{Samples from TA2I-edit - Page 1}
\end{figure*}

\begin{figure*}[!htbp]
    \centering
    \captionsetup{width=\textwidth}
    \includegraphics[page=2, width=\linewidth]{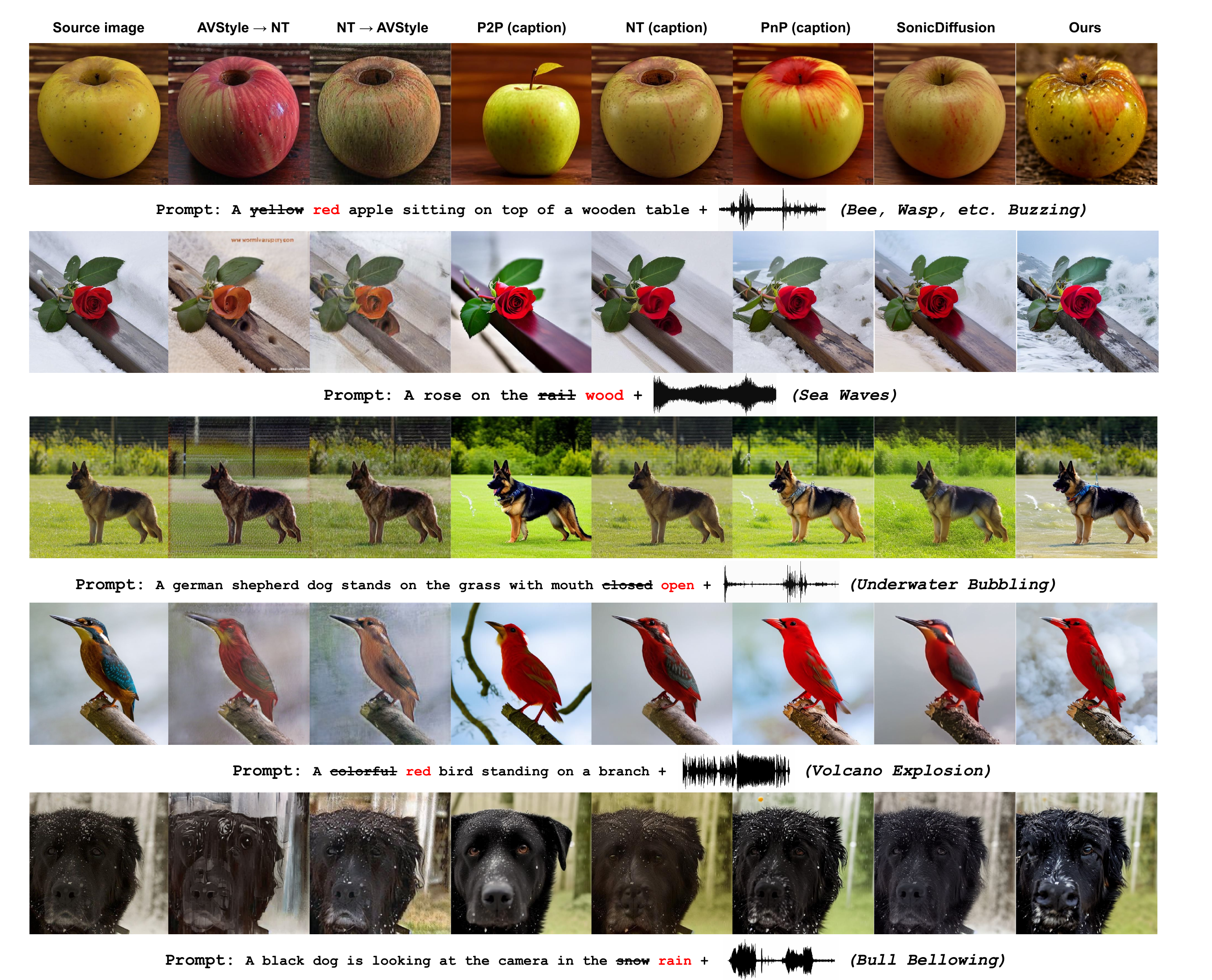}
    \caption{Samples from TA2I-edit - Page 2}
\end{figure*}

\clearpage
\section{Further Samples on A2V-edit}\label{sec:sup_a2v}

\begin{figure*}[!htbp]
    \centering
    \captionsetup{width=\textwidth}
    \includegraphics[page=1, width=\linewidth]{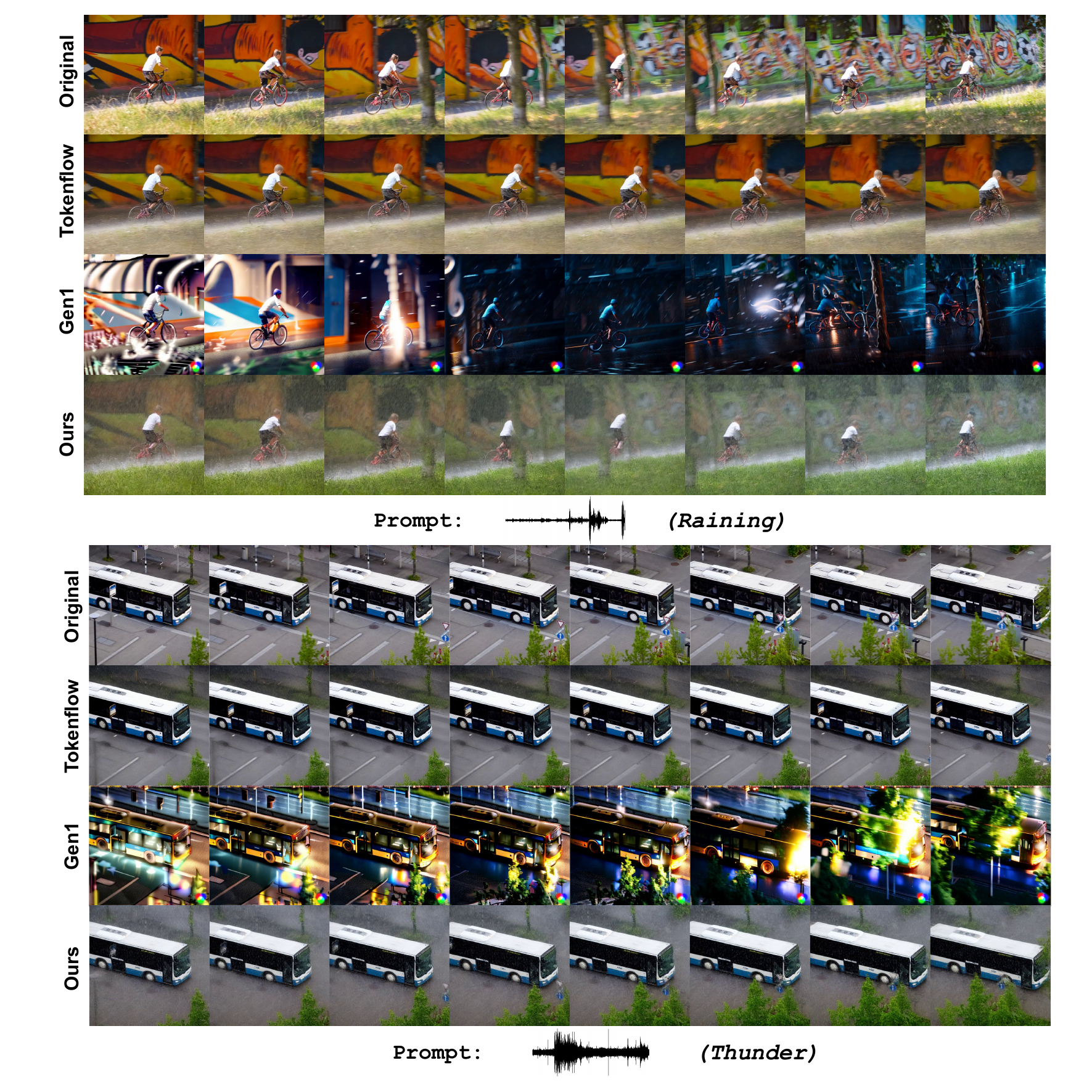}
    \caption{Samples from A2V-edit - Page 1}
\end{figure*}

\begin{figure*}[!htbp]
    \centering
    \captionsetup{width=\textwidth}
    \includegraphics[page=2, width=\linewidth]{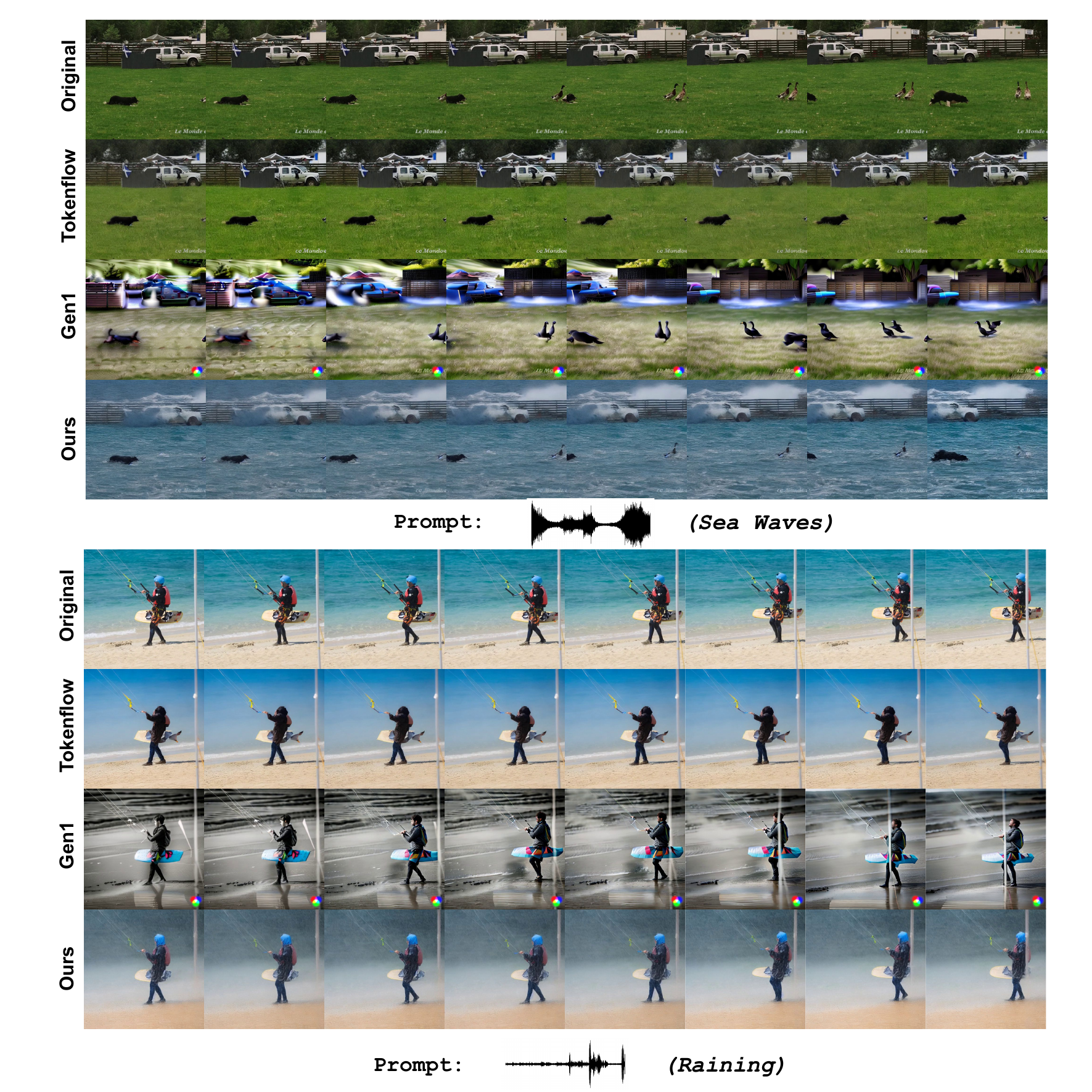}
    \caption{Samples from A2V-edit - Page 2}
\end{figure*}

\clearpage
\section{Further Samples on TA2V-edit}\label{sec:sup_ta2v}

\begin{figure*}[!htbp]
    \centering
    \captionsetup{width=\textwidth}
    \includegraphics[page=1, width=\linewidth]{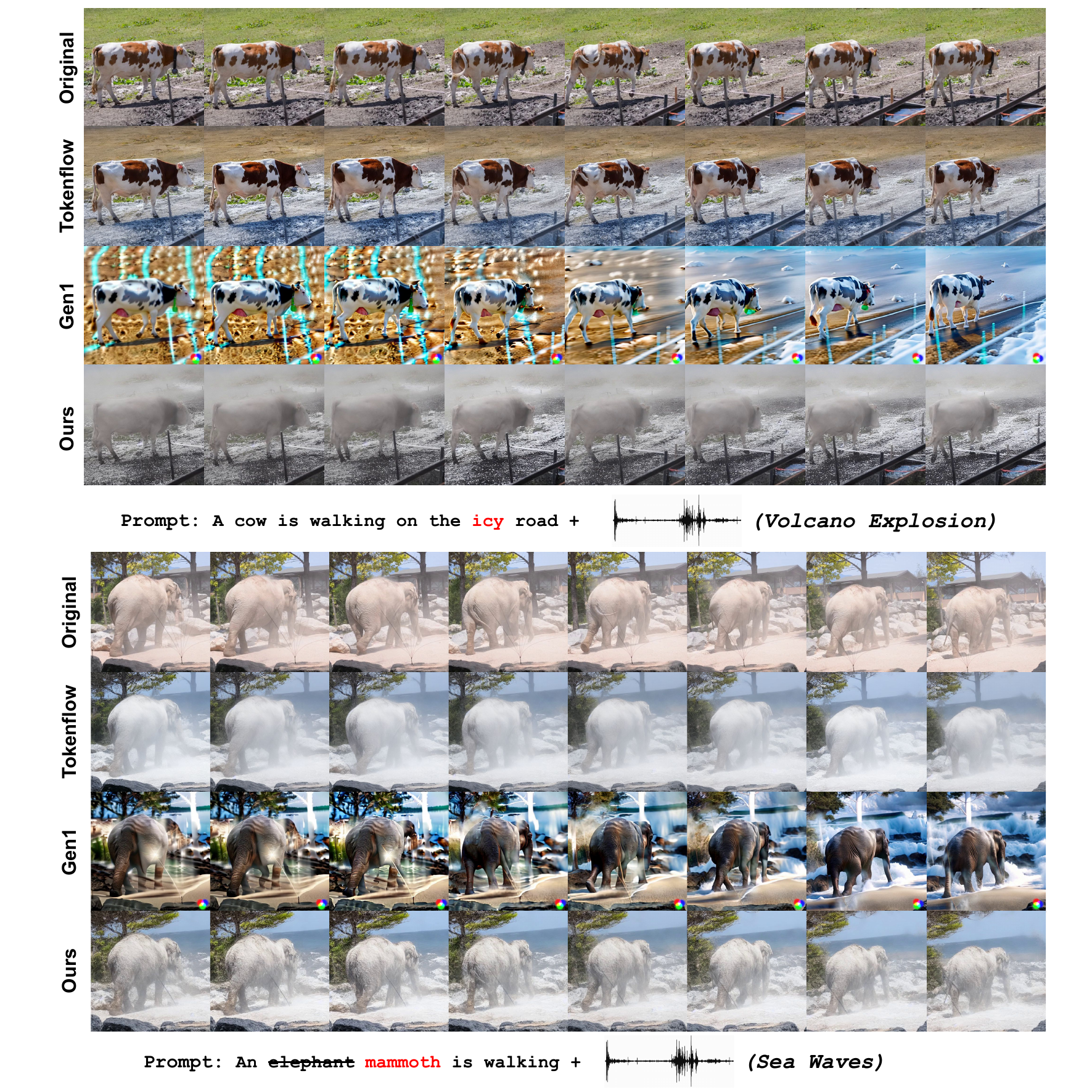}
    \caption{Samples from TA2V-edit - Page 1}
\end{figure*}

\clearpage
\section{Detailed Evaluation results on TA2I-edit}\label{sec:sup_cascaded}

\begin{table*}[ht]
    \centering
    \captionsetup{width=\linewidth}
    \caption{
        Evaluation results for Audio-guided Image Editing (A2I-edit) and Text and Audio-guided Image Editing (TA2I-edit). 
        \textbf{Str. Dist} represents Structure Distance. 
        \textbf{CLIP\_Audio} is the CLIP score using the audio editing prompt, 
        \textbf{CLIP\_Text} is the CLIP score using the text editing prompt, and 
        \textbf{CLIP\_integrated} is the CLIP score using a combined text and audio caption. 
    }
    \label{tab:image_edit_results_full}
    \resizebox{\linewidth}{!}{%
    \begin{tabular}{llcccccc}
        \toprule
        && \textbf{Str. Dist$(\downarrow)$} & \textbf{LPIPS$(\downarrow)$} & \textbf{CLIP\_Audio$(\uparrow)$} & \textbf{CLIP\_Text$(\uparrow)$} & \textbf{CLIP\_integrated$(\uparrow)$} \\
        \midrule
        \multirow{7}{*}{\textbf{TA2I-edit}} 
            & AVStyle $\rightarrow$ P2P & 0.0932 & 0.4400 & 12.3686 & 26.1706 & 21.6530 \\
            & AVStyle $\rightarrow$ NT & 0.0668 & 0.4174 & 13.1312 & 25.2388 & 21.1613 \\
            & AVStyle $\rightarrow$ PnP & 0.0675 & 0.3994  & 12.6587 & 25.5583 & 21.3918 \\

            & Sonic Diffusion $\rightarrow$ P2P & 0.0864 & 0.4164 & 12.6119 & 26.1812 & 21.7774 \\
            & Sonic Diffusion $\rightarrow$ NT & 0.0616 & 0.3632 & 13.1615 & 25.4673 & 21.7049 \\
            & Sonic Diffusion $\rightarrow$ PnP & 0.0628 & 0.3644 & 12.8310 & \textbf{26.2953} & 22.1569 \\

            & P2P $\rightarrow$ AVStyle & 0.0529 & 0.2688 & 14.5720 & 26.0751 & 22.1799 \\
            & P2P $\rightarrow$ Sonic Diffusion & 0.0202 & 0.1207 & 13.1563 & 25.2779 & 17.7859 \\
            
            & NT $\rightarrow$ AVStyle & 0.0369 & 0.2508 & 14.4867 & 25.5348 & 21.6772 \\
            & NT $\rightarrow$ Sonic Diffusion & 0.01829 & 0.1443 & 13.1250 & 24.5920 & 20.8667 \\

            & PnP $\rightarrow$ AVStyle & 0.0172 & 0.2757 & 14.5875 & 25.8681 & 22.0208 \\
            & PnP $\rightarrow$ Sonic Diffusion & \textbf{0.0159} & \textbf{0.1199} & 13.0562 & 25.0826 & 21.2290 \\
            
            & P2P (caption) & 0.0591 & 0.3278 & 12.3059 & 24.9569 & 21.1932 \\
            & NT (caption) & 0.0192 & 0.1802 & 12.8379 & 24.3057 & 21.6057 \\
            & PnP (caption) & 0.0264 & 0.2108 & 13.0695 & 24.9375 & 21.3103 \\
            
            & Sonic Diffusion & 0.0179 & 0.1551 & 12.9011 & 23.6654 & 20.1574 \\
            
            & \textbf{Ours (mean)} & 0.0204 & 0.1729 & 14.0025 & 24.2257 & 20.1574 \\
            & \textbf{Ours} & 0.0333 & 0.2392 & \textbf{15.0699} & 24.1283 & \textbf{23.0356} \\
        \bottomrule
    \end{tabular}
    }
\end{table*}

\clearpage
\section{Further Samples with multiple editing prompts}\label{sec:sup_multi}

\begin{figure*}[!htbp]
    \centering
    \captionsetup{width=\textwidth}
    \includegraphics[page=1, width=0.6\linewidth]{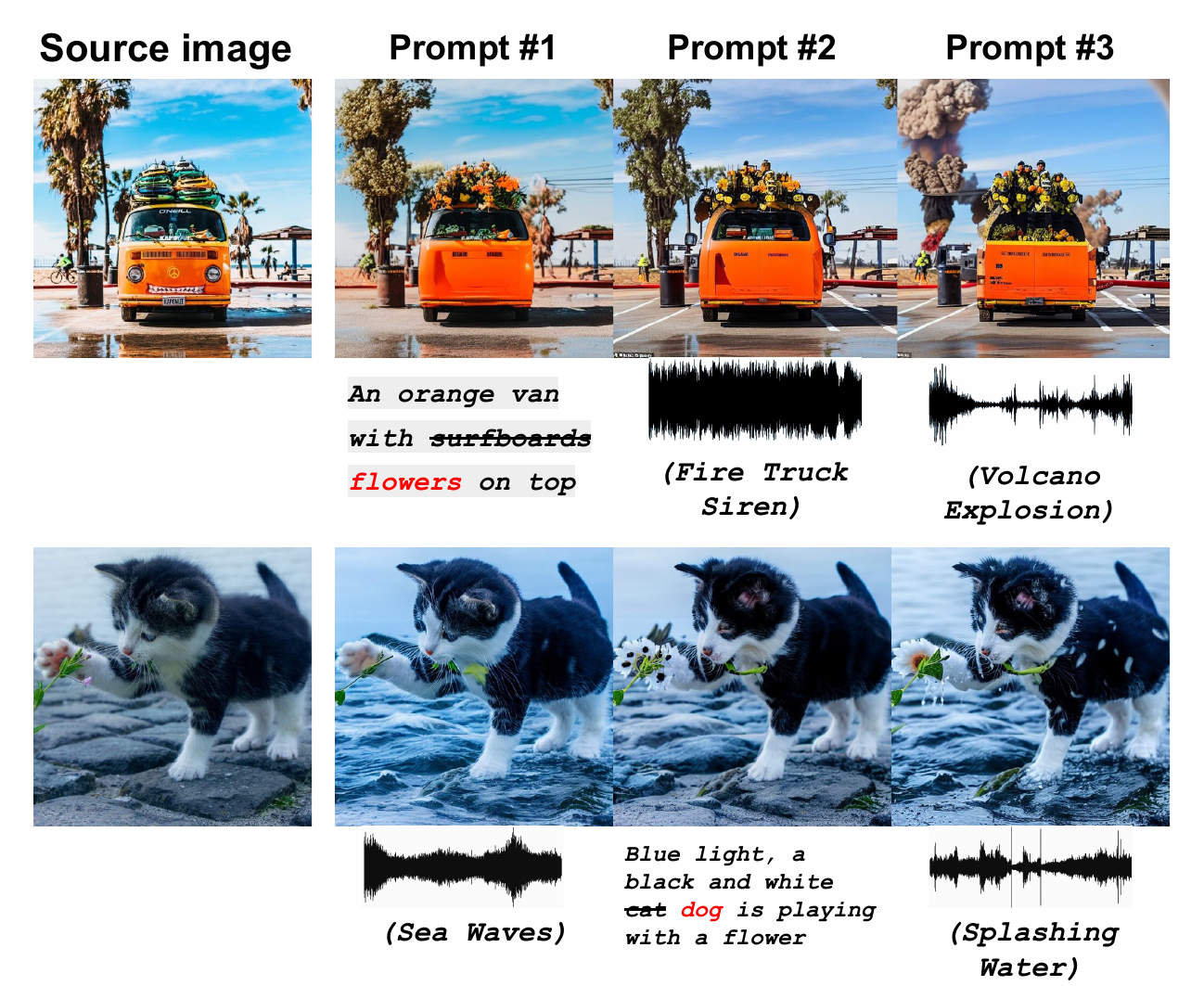}
    \caption{Samples with multiple editing prompts - Page 1}
\end{figure*}

\begin{figure*}[!htbp]
    \centering
    \captionsetup{width=\textwidth}
    \includegraphics[page=2, width=0.6\linewidth]{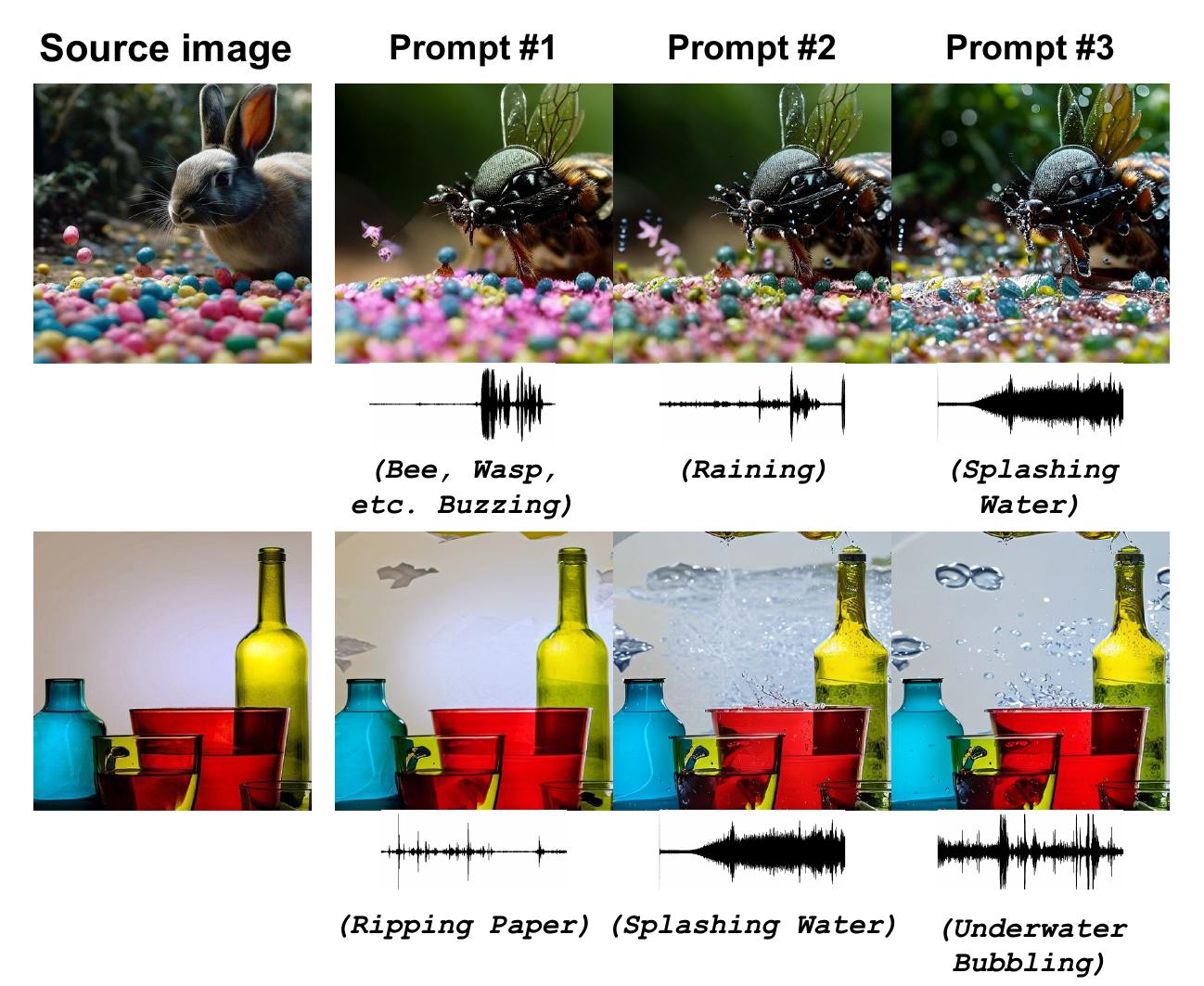}
    \caption{Samples with multiple editing prompts - Page 2}
\end{figure*}